\documentclass[11pt]{article}
\usepackage[final]{acl}
\usepackage{times}
\usepackage{latexsym}
\usepackage{svg}
\usepackage{multirow}
\usepackage{supertabular} 
\usepackage{booktabs}
\usepackage{lscape}
\usepackage{longtable}
\usepackage{lipsum}
\usepackage{hyperref}
\usepackage{siunitx}
\usepackage[T1]{fontenc}

\usepackage{graphicx}
\usepackage{ifthen}
\usepackage{amssymb}
\usepackage{microtype}

\usepackage{inconsolata}

\usepackage{graphicx}
\usepackage{tabularray}
\usepackage{adjustbox}
\usepackage{soul}

\usepackage[most]{tcolorbox}

\tcbset {
  base/.style={
    arc=0mm, 
    bottomtitle=0.5mm,
    boxrule=0mm,
    colbacktitle=black!10!white, 
    coltitle=black, 
    fonttitle=\bfseries, 
    left=2.5mm,
    leftrule=1mm,
    right=3.5mm,
    title={#1},
    toptitle=0.75mm, 
  }
}

\definecolor{brandblue}{rgb}{0.34, 0.7, 1}
\newtcolorbox{mainbox}[1]{
  colframe=brandblue, 
  base={#1}
}

\newtcolorbox{subbox}[1]{
  colframe=black!30!white,
  base={#1}
}
\title{Is Small Language Model the Silver Bullet to Low-Resource Languages Machine Translation?}

\author{
 \textbf{Yewei Song\textsuperscript{$\spadesuit$1}},
 \textbf{Lujun Li\textsuperscript{$\spadesuit$1}},
 \textbf{Cedric Lothritz\textsuperscript{2}},
 \\
 \textbf{Saad Ezzini\textsuperscript{3}},
  \textbf{Lama Sleem\textsuperscript{1}},
 \textbf{Niccolo' Gentile\textsuperscript{4}},
 \\
 \textbf{Radu State\textsuperscript{1}},
 \textbf{Tegawendé F. Bissyandé\textsuperscript{1}},
 \textbf{Jacques Klein\textsuperscript{1}},
\\
 \textsuperscript{1}University of Luxembourg,
 \textsuperscript{2}Luxembourg Institute of Science and Technology,
 \\
 \textsuperscript{3}King Fahd University of Petroleum and Minerals,
 \textsuperscript{4}Foyer S.A.,
\\
 \small{
   \textbf{Correspondence:} \href{mailto:email@domain}{yewei.song@uni.lu}
 }
}

\newboolean{showcomments}
\setboolean{showcomments}{true}
\ifthenelse{\boolean{showcomments}}
 { \newcommand{\mynote}[2]{
      \fbox{\bfseries\sffamily\scriptsize#1}
        {\small$\blacktriangleright$\textsf{\emph{#2}}$\blacktriangleleft$}}}
        { \newcommand{\mynote}[2]{}}

\begin{document}
\maketitle
\def\thefootnote{$\spadesuit$}\footnotetext{These authors contributed equally to this work.}\def\thefootnote{\arabic{footnote}}

\begin{abstract}
Low-resource languages (LRLs) lack sufficient linguistic resources and are underrepresented in benchmark datasets, resulting in persistently lower translation quality than high-resource languages, especially in privacy-sensitive and resource-limited contexts. Firstly, this study systematically evaluates state-of-the-art smaller Large Language Models in 200 languages using the FLORES-200 benchmark, highlighting persistent deficiencies and disparities in the translation of LRLs. To mitigate these limitations, we investigate knowledge distillation from large pre-trained teacher models to Small Language Models (SLMs) through supervised fine-tuning. The results show substantial improvements; for example, the translation performance of English$\rightarrow$ Luxembourgish (EN$\rightarrow$ LB), measured by the LLM-as-a-Judge score, increases from 0.36 to 0.89 in the validation set for Llama-3.2-3B. We further investigate various fine-tuning configurations and tasks to clarify the trade-offs between data scale and training efficiency, verify that the model retains its general capabilities without significant catastrophic forgetting after training, and explore the distillation benefits to other LRLs on SLMs (Khasi, Assamese, and Ukrainian). In general, this work exposes the limitations and fairness issues of current SLMs in LRL translation and systematically explores the potential of using the distillation of knowledge from large to small models, offering practical, empirically grounded recommendations to improve LRL translation systems\footnote{\scriptsize \url{https://anonymous.4open.science/r/mt_luxembourgish-408D}}.
\end{abstract}

\section{Introduction}

Low-resource languages (LRLs) suffer from a lack of critical linguistic resources, and this scarcity is often rooted in socioeconomic, geographical, and political factors, which contribute to their poor support in academic research and industrial applications \citep{nigatu2024zenos}.

Recent progress has greatly improved translation for \textbf{H}igh-\textbf{R}esource \textbf{L}anguages (HRLs), large performance gaps remain for LRLs, especially in areas like finance and government, where privacy is crucial and models often run on low-power, offline devices \cite{zhong2024opportunitieschallengeslargelanguage}. Recent multilingual transfer and pretraining learning methods \citep{conneau2019unsupervised, artetxe2019massively}, exemplified by initiatives such as No Language Left Behind (NLLB; \citealp{nllb2022no}), have greatly improved cross-lingual representation and translation quality. However, these approaches typically rely on high quality substantial parallel datasets, which are rarely available to LRLs, especially in formal domains such as news and official communications. Translation from LRLs into HRLs is typically more straightforward because of the greater abundance of target-side resources, while the reverse direction remains considerably challenging. Furthermore, portable \textbf{S}mall \textbf{L}anguage \textbf{M}odels(SLMs), which is of a parameter size less than 4B, critical for mobile devices, exhibits weaker performance in LRL tasks, exacerbating the existing translation gap. To explore the current ``landscape'' and the applicability of transformer-based models for LRLs, this article makes the following three key contributions. 


\textbf{First}, we quantitatively analyze LLM performance in 200 languages using the FLORES-200 benchmark (Section~4), highlighting disparities that affect underrepresented languages. Our analysis underscores the concerning state of LLMs for LRLs and reveals that SLMs exhibit even more pronounced deficiencies in these languages. \textbf{Second}, we demonstrate that alternative single-sided data sources (e.g., news articles and monolingual resources) can be used to distill knowledge from teacher models (Section~5), improving translation quality on both sides (\text{LRL}~$\Leftrightarrow$~\text{HRL}) in SLMs and help address the scarcity of parallel data. \textbf{Third}, we investigate a range of fine-tuning configurations and methodologies (Section~6), offering practical guidelines for Supervised Fine-Tuning (SFT) utilizing distilled data generated by teacher models.


\section{Related Work}

\subsection{Generative Models}

Transformer-based architectures have significantly advanced machine translation through multilingual embeddings and nuanced language generation \citep{zhao2023transformer,zhao2024largelanguagemodelshandle}. Current translation models typically employ encoder-decoder architectures with attention mechanisms \citep{bahdanau2014neural,vaswani2017attention,naveed2024comprehensiveoverviewlargelanguage}, or decoder-only frameworks exemplified by the GPT series, recognized for computational efficiency and ease of fine-tuning \citep{gao2022encoderdecoderredundantneuralmachine,hendy2023goodgptmodelsmachine}. Recent methods such as back-translation \citep{sennrich2016improving}, unsupervised translation \citep{lample2018unsupervised}, and multilingual systems such as OPUSMT \citep{tiedemann2020opus} further enhance translation quality. However, decoder-only models often face limitations for LRLs due to predominantly English-centric training data \citep{DBLP:journals/corr/abs-2005-14165,hasan2024largelanguagemodelsspeak}, leading to translation inaccuracies and hallucinations \citep{benkirane2024machinetranslationhallucinationdetection}. Despite these challenges, recent findings suggest that decoder-only architectures may outperform encoder-decoder models in certain translation tasks \citep{gao2022encoderdecoderredundantneuralmachine,silva-etal-2024-benchmarking}, motivating our investigation into their application for improving LRL translations.

\subsection{Limited Support for LRLs}

Despite considerable advances, current LLMs offer insufficient support for low-resource languages. Research consistently demonstrates substantial performance degradation in LRL translation tasks compared to high-resource languages \citep{robinson2023chatgpt}. This performance gap arises primarily from unbalanced training datasets that overwhelmingly favor high-resource languages \citep{blasi2021systematic,lankford2021transformers}. Furthermore, tokenization biases and uneven data exposure hinder the ability of the models to accurately capture linguistic nuances unique to LRLs \citep{shen2024language}. Addressing these shortcomings requires targeted data enhancement techniques and customized fine-tuning methods to significantly enhance LLM capabilities for low-resource language tasks \citep{elsner2024shortcomings,li2024language}.

\section{Research Questions}

\textbf{RQ1:} How effectively do large language models (LLMs) address low-resource machine translation, and what are the comparative performance gaps in translation quality among different model sizes and languages?\\

\noindent \textbf{RQ2:} How much can distillation from monolingual LRL-side data improve the translation performance of smaller LLMs in low-resource languages?\\

\noindent \textbf{RQ3:} How do different SFT settings affect model performance on low-resource language translation tasks and do they risk compromising the model’s general capabilities? Are similar improvements observed in diverse LRLs?\\

\section{Investigation of LRLs}
\label{sec:lrl}

\label{sec:lrlllm}
\subsection{Situation of Language Support}

Recent investigations have revealed that although LLMs are increasingly advertised as multilingual, their effective support in languages is often limited to a subset of HRLs. Moreover, systematic evaluations of language-specific performance remain scarce (for example \cite{lai2024llms, marchisio2024understanding, lifewire2024llama3, ahuja2023megaverse}). Table~\ref{tab:llm_support} summarizes several models included in our experiments, their approximate parameter sizes, and the estimated number of languages they reportedly support. These figures are derived from official model documentation, benchmarking reports, and recent academic studies.

\begin{table}[!htbp]
\centering

\begin{adjustbox}{width=0.47\textwidth}
    \setlength{\tabcolsep}{4pt}
    \begin{tabular}[t]{@{}llcc@{}}
    \toprule
    \textbf{Model} & \textbf{Size} & \textbf{Languages} & \textbf{Date} \\
    \midrule
    GPT-4o-mini                  & ---   & $\sim$25 & Jul. 2024 \\
    Llama-3.1-8B-it              & 8B/3B    & $\sim$30 & Jul. 2024 \\
    Llama-3.2-3B-it              & 3B    & $\sim$20 & Sept. 2024 \\
    Mistral-8B-Instruct-2410     & 8B    & $\sim$25 & Oct. 2024 \\
    Phi-3-mini-4k-instruct       & 4B    & $\sim$20 & Apr. 2024 \\
    Phi-3.5-mini-instruct        & 4B    & $\sim$20 & Aug. 2024 \\
    Qwen2.5 Instruct        & 1.5B/3B  & $\sim$25 & Sept. 2024 \\
    Gemma-2 Instruct                & 2B/9B    & $\sim$20 & Jul. 2024 \\
    \bottomrule
    \end{tabular}
\end{adjustbox}
\caption{Overview of Multilingual Support in LLMs}
\label{tab:llm_support}
\vspace{-0.8em}
\end{table}

Despite these encouraging multilingual claims, the existing literature reveals that rigorous language-specific performance evaluations, especially for low-resource languages, are insufficient. Most current research focuses on high-resource benchmarks, leaving open critical questions about fairness and the accessibility of LLMs for diverse linguistic communities.

\subsection{Evaluating Language Ability}

\begin{figure}[htbp]
    \centering
    \includegraphics[width=\linewidth]{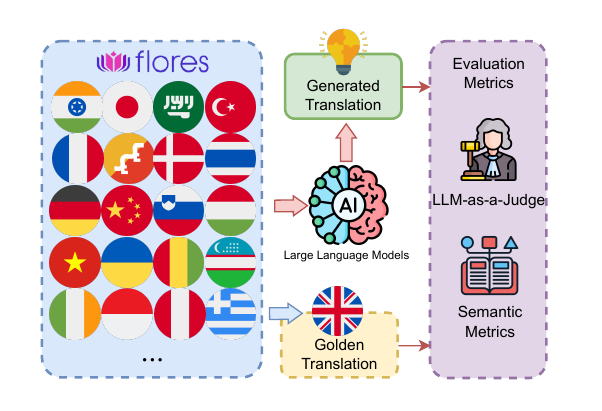}
    \caption{Evaluation pipeline on FLORES-200}
    \label{fig:llmeval}
    \vspace{-1em}
\end{figure}

We use the \textbf{FLORES-200} benchmark to systematically assess the performance of LLMs in multilingual machine translation tasks~\cite{nllb2022no,flores101,nedlguzm}. FLORES-200 offers rigorously curated human-validated translation datasets across 200 languages that span diverse linguistic families and writing systems, making it highly effective for evaluating translation quality in high-resource and low-resource linguistic contexts. Our experiments leverage the full FLORES-200 dataset to comprehensively evaluate translation quality across as many languages as possible, emphasizing translations from various source languages into English.

In addition to traditional metrics, we evaluated translation quality using the \textbf{LLM}-\textbf{A}s-\textbf{A}-\textbf{J}udge (LLMaaJ) scores~\cite{niklaus2025swiltrabenchswisslegaltranslation}, which uses a large LLM to score translations from 0 to 1 based on semantic equivalence and naturalness. A score of 1.0 denotes a perfect translation and 0.0 a totally incorrect one. In practice, we consider a score $\geq 0.8$ as indicative of a good translation. Research has shown that LLMaaJ tolerates synonyms, paraphrases, and cross-linguistic structural variations, enabling it to better assess translation quality when there are multiple valid phrasings or when grammatical and typological differences (e.g., omitted pronouns) are acceptable\cite{zheng2023judging, piergentili2025llm}.



Regarding the LLMs investigated, as shown in Figure \ref{fig:llmeval}, we systematically traversed prominent proprietary APIs and open source models (refer to Table~\ref{tab:llm_support}), presenting results using LLMaaJ metrics with quantitative semantic evaluations. Detailed LLMaaJ and BLEU scores for all source-to-English translations are provided in the Appendix Table~\ref{tab:LLMAAJALL} and the Appendix Table~\ref{tab:BLEU}.


\subsection{LLMs performance in FLORES-200}

\begin{figure*}[htbp]
    \vspace{-1em}
    \centering
    \includegraphics[width=\linewidth]{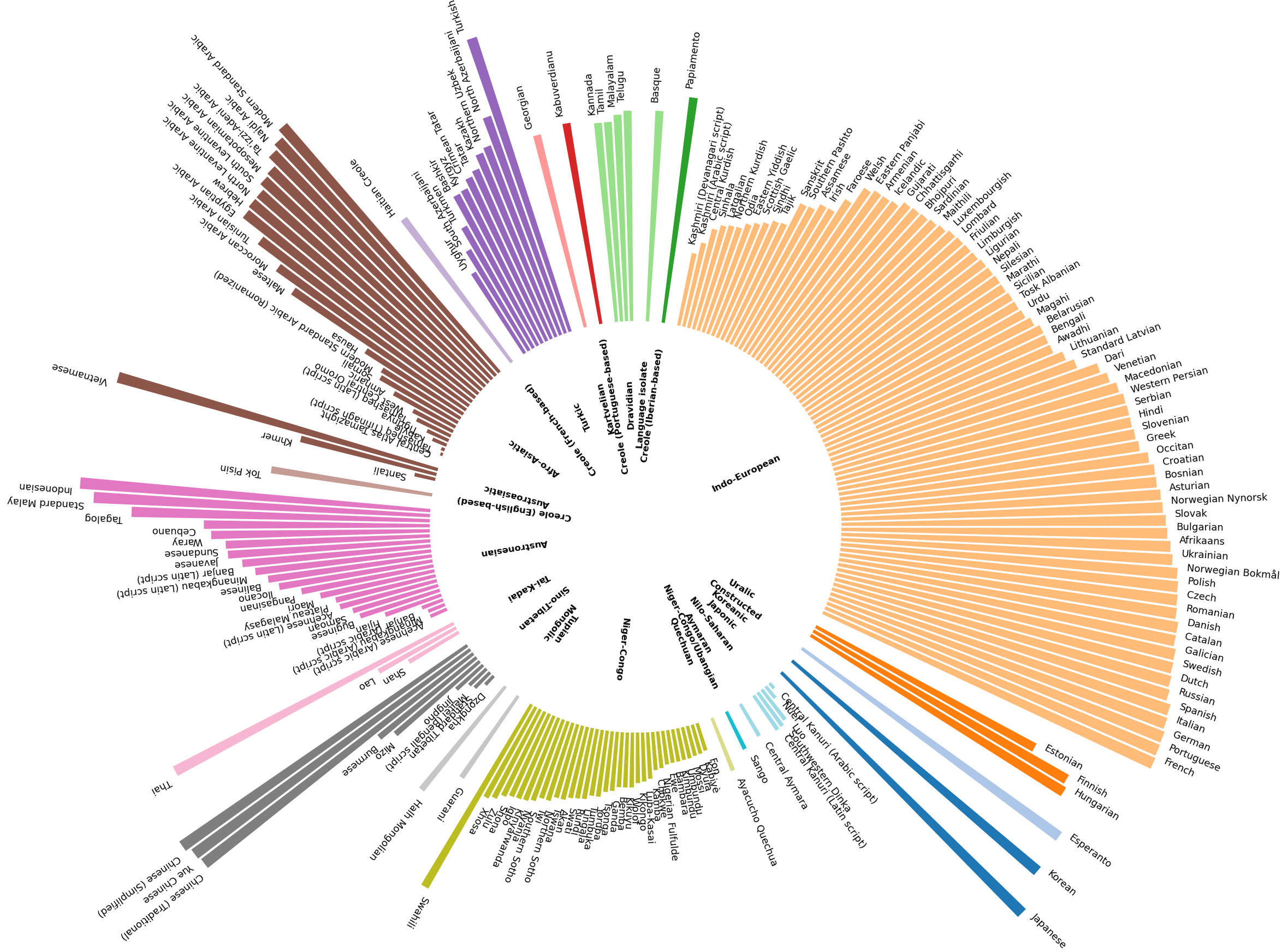}
    \caption{"Low-Resource" Linguistic results grouped by language families}
    \label{fig:circleplot}
    \vspace{-1em}
\end{figure*}
In this paper, we replace the regional map with Figure~\ref{fig:circleplot}, which more precisely visualizes the distribution of languages in our set of evaluations by linguistic family and script, thus answering RQ1. Each bar length is calculated based on the average score, explicitly excluding the GPT4o‑mini model's score. This family-level visualization makes it easier to identify which LRLs are included in our experiments and how they are situated in the broader typological space.

Each bar in Figure~\ref{fig:circleplot} represents one language, grouped by its primary family, with bar length corresponding to the average LLMaaJ score. The figure reveals that LRLs are not evenly distributed across families: many under-resourced African, Austronesian, and Indigenous American languages cluster toward the lower end of the performance spectrum, while certain Indo-European LRLs (e.g., Luxembourgish, Maltese) perform moderately better, likely due to greater data availability or proximity to high-resource relatives.

The circular layout also highlights structural gaps in the evaluation set. Languages absent from FLORES-200—such as many North American Indigenous languages—do not appear here, not because models perform well on them, but because no evaluation data exists. This is particularly relevant for languages with small speaker populations or those concentrated in politically marginalized communities, which remain invisible in current multilingual benchmarks.

Consistent with previous work~\cite{nekoto-etal-2020-participatory,joshi-etal-2020-state}, the lowest scores are observed for many Niger–Congo, Austronesian, and smaller Afro-Asiatic languages, reflecting the severe data scarcity. In contrast, LRLs in Eastern Europe and South/Southeast Asia—such as Macedonian or Sinhala—achieve slightly higher average scores, possibly benefiting from historical ties to better-supported high-resource languages. However, the overall pattern remains unchanged: LRLs across all families systematically lag behind high-resource languages, underscoring the need for targeted data collection, typologically diverse benchmarks, and bias mitigation strategies to ensure equitable progress in multilingual NLP.

\subsection{Gap between Dwarf(Smaller) and Giant LLMs}
\begin{figure*}[htbp]
  \tabcolsep=0pt
  \begin{tabular*}{\textwidth}{@{\extracolsep{\fill}}ccc}
    \includegraphics[width=0.33\textwidth]{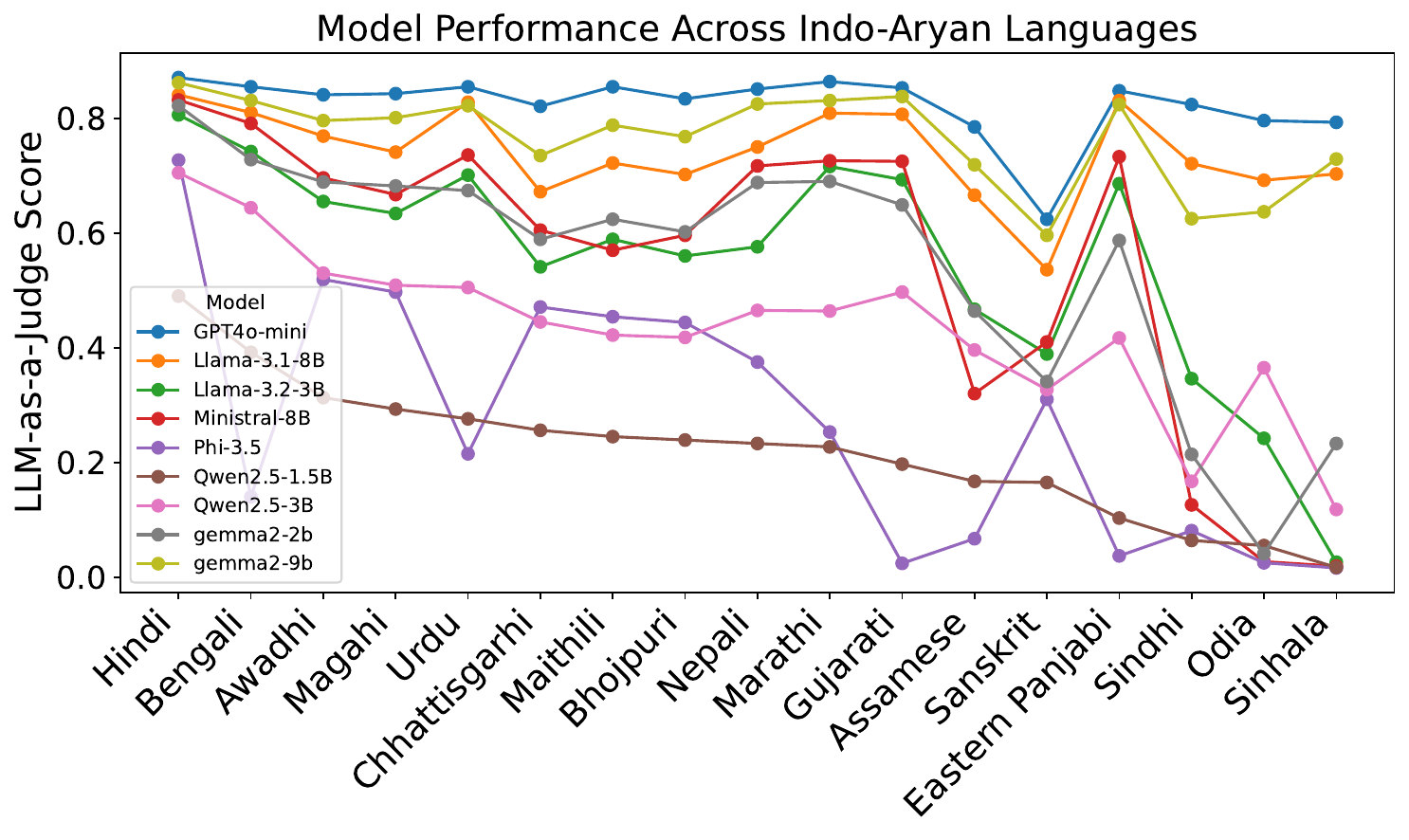} &
    \includegraphics[width=0.33\textwidth]{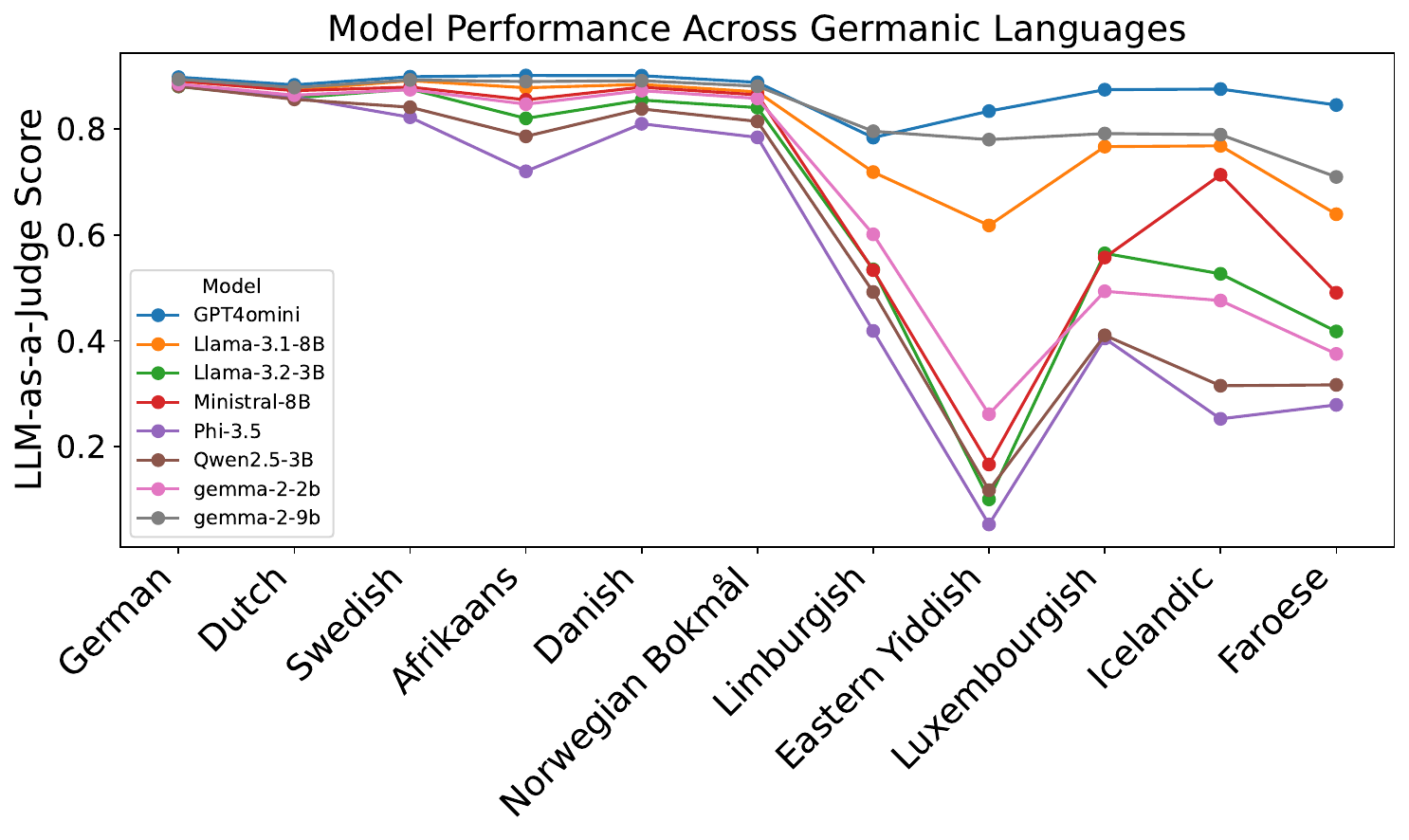} &
    \includegraphics[width=0.33\textwidth]{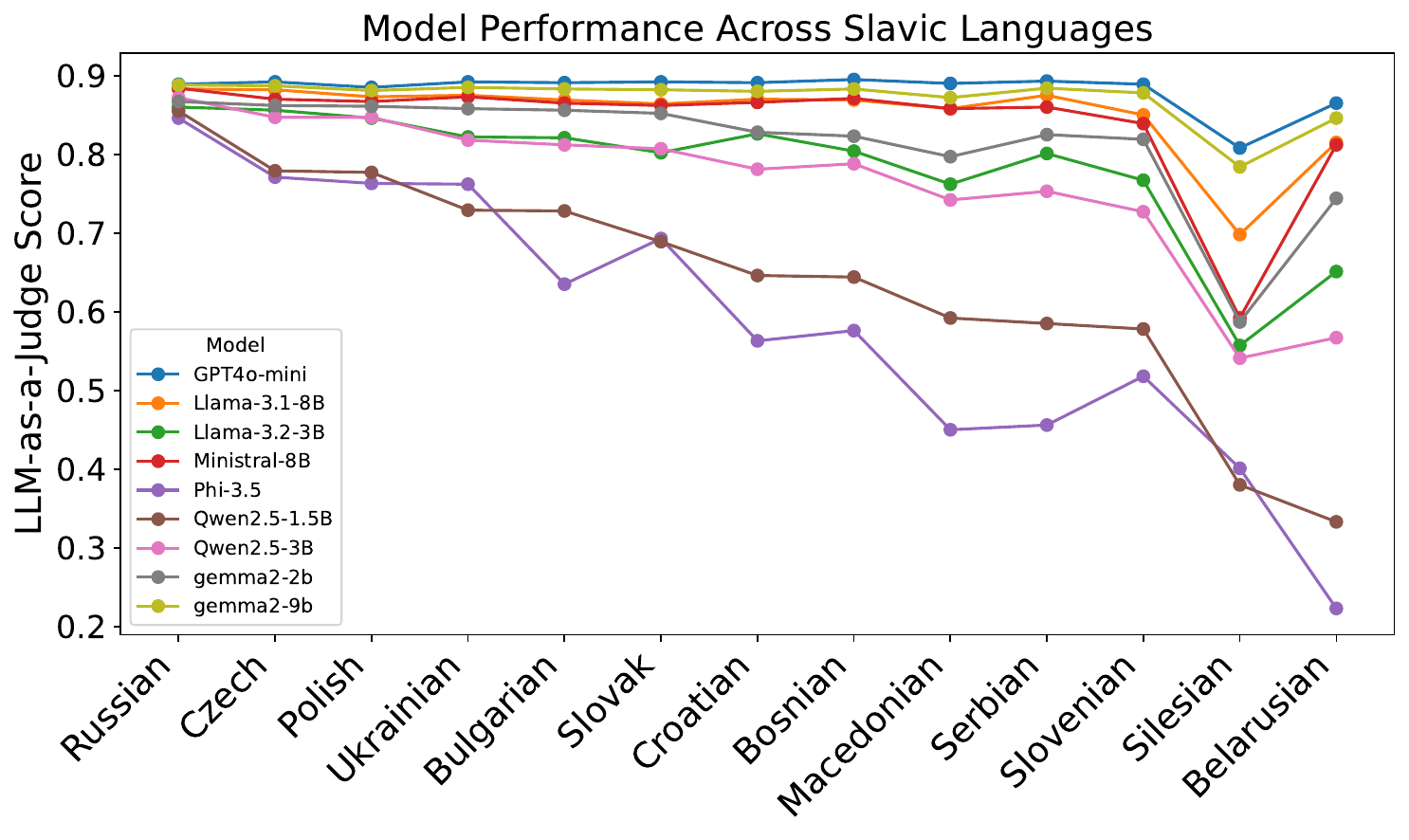} \\
    \small (a) Indo-Aryan &
    \small (b) Germanic &
    \small (c) Slavic 
  \end{tabular*}
\caption{LLMaaJ scores of smaller LLMs on Indo-Aryan/Germanic/Slavic languages for to-English translation}
\label{fig:germanic}
\vspace{-1em}
\end{figure*}

Across the Indo-Aryan, Germanic, and Slavic branches in Figure~\ref{fig:germanic} (panels (a)--(c)), we observe a consistent pattern: \textbf{smaller LLMs suffer a substantially larger performance drop on low-resource languages (LRLs)} than on high-resource ones, while larger LLMs degrade far less. Concretely, LRLs such as Sinhala (Indo-Aryan), Luxembourgish (Germanic), and Silesian (Slavic) exhibit steep declines in smaller models but remain comparatively competitive in larger models, as visualized in Figure~\ref{fig:germanic}. This disparity indicates a systematic bias in current systems—particularly pronounced in smaller models—toward high-resource languages.

Addressing this gap calls for improved LRL data curation, transfer via knowledge distillation from larger LLMs, and more inclusive evaluation suites that foreground LRLs, alongside bias-mitigation strategies, so advances in NLP benefit all language communities. In principle, according to the Universal Approximation Theorem \cite{HORNIK1991251}, if we assume that the translation task can be modeled as a linear mapping from one semantic space to another, then a small-capacity network is limited in its ability to capture complex patterns and is therefore more susceptible to being perturbed or contaminated by the large-volume data typical of HRLs. Consequently, fine-tuning becomes intuitively all the more important for smaller models.

\section{Fine-tuning on LRLs}

\subsection{Background and language selection}

As highlighted in the previous section, several low-resource languages, such as Luxembourgish and Assamese (Figure~\ref{fig:germanic}), show a substantial translation quality gap among between large and small models. In this article, Luxembourgish serves as a representative case. Although officially recognized, it lacks sufficient high-quality corpora resources, leading to poor performance in SLMs. Its blend of Germanic roots and French influence adds complexity to NLP tasks. While larger LLMs handle Germanic languages reasonably well, they struggle with LRLs like Luxembourgish. Previous efforts to address this include LuxemBERT \citep{lothritz2022luxembert}, LuxT5 \citep{plum2024text}, and LetzTranslate \citep{song2023letztranslatelowresourcemachine}, a low-resource translation system based on OPUS-MT.

To examine generalizability, we additionally include Ukrainian, Assamese and Khasi (an endangered language), both exhibiting similar linguistic and resource profiles, as supplementary tasks to broaden the scope of the analysis. Furthermore, generating LRL from English is more challenging for LLMs than in the reverse direction of previous research~\cite{howcroft2022most}. Regarding translation performance, LLMs exhibit a certain degree of fluent translation from LRL to English, but not vice versa~\cite{gao2020improving}. This asymmetry is also reflected to some extent in the hallucination issues observed when generating Luxembourgish, more details can be found in the appendix \ref{apd:repetionproblems}.

\subsection{Distillations and Soft-Target Quality}


In our scenario, having only a Luxembourgish corpus without English translations rules out conventional parallel-corpus training approaches, accurately reflecting the typical data situation and model generation of LRLs. To bridge the gap between comprehension and generation in this low-resource scenario, we propose a distillation-based approach. Using a teacher model that demonstrates a robust understanding of Luxembourgish, we can distill its knowledge into a student model using the available LRL single-side corpus. This process is expected to enhance the generation capabilities of the student model, enabling it to produce high-quality Luxembourgish output despite the limited data, and thus address the core challenge of low-resource language translation. According to further human labeling of our GPT-4o distillation dataset in Luxembourgish to English translation, \textbf{92\%} of our samples were marked as fully correct.

\subsection{Data Collection and Augmentations}
\label{subsection: Data Collection}

For the training data set, we constructed a Luxembourg data set using multiple sources, including the LuxemBERT corpus, example sentences in the Luxembourg Online Dictionary (LOD) dataset\footnote{https://data.public.lu/en/datasets/letzebuerger-online-dictionnaire-lod-linguistesch-daten/}, and additional news articles collected from previous research published data on RTL Ltzebuerg\footnote{https://www.rtl.lu/}, following the LuxemBERT work.

Previous research has demonstrated that integrating dictionary entries can effectively enrich low-resource translation systems by providing explicit lexical alignments and clarifying semantic nuances. For example, Ghazvininejad's work improved translation fidelity in settings where parallel data is scarce \cite{ghazvininejad2023dictionary}. Inspired by these findings, we also explore how the addition group of datasets with dictionary checks using LOD can complement our distillation approach as shown in Figure \ref{fig:pipeofaug}. Details of using the dictionary usage in the Appendix \ref{apd:dictrag}.
\begin{figure}[!htbp]
    \includegraphics[width=0.53\textwidth]{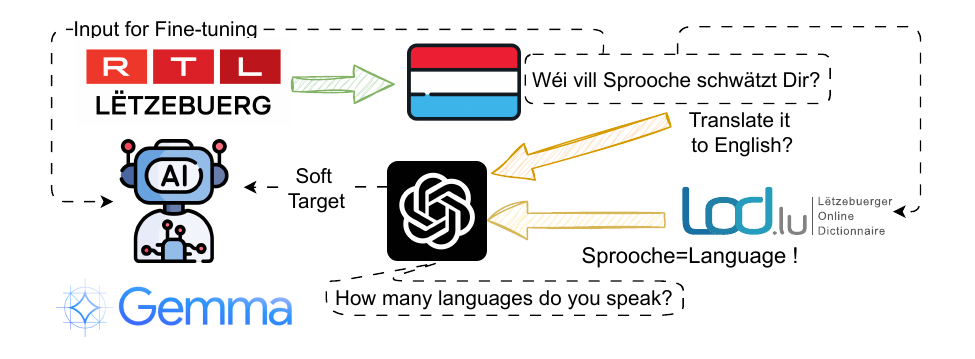}
    \caption{Pipeline of data augmentation}
    \label{fig:pipeofaug}
    \vspace{-1em}
\end{figure}

\subsection{Model SFT}

State-of-the-art decoder-only models are typically trained in three stages: pre-training, Supervised Fine-Tuning (SFT) and further tuning using Reinforcement Learning from Human Feedback \cite{ouyang2022traininglanguagemodelsfollow}. When combined with techniques such as  Low-Rank Adaptation (LoRA: \cite{hu2022lora}), SFT can also significantly improve performance on tasks with fewer resources. In this study, in order to validate our distillation strategy, we primarily adopt model distillation from LRL-side corpora and then incorporate SFT to equip the model to generate and reverse-translate the target language. We use the classical approach of supervised instruction fine-tuning for two different models. The fundamental logic is to provide the model with input prompts and corresponding responses, optimizing the model to minimize prediction loss within this fixed framework. In decoder-only models, text generation is performed recursively by predicting the probability distribution over the vocabulary for the next token. SFT aims primarily to maximize the probability of the correct next token, thereby teaching the trained model the relationships between semantics, vocabulary, and syntax in LRL, as well as their correspondence with HRL.

\section{Experiments}
\subsection{Models and Datasets}
\begin{table*}[htbp]
\centering
\resizebox{\textwidth}{!}{
\begin{tblr}{
  rowsep=0.9pt,
  row{1} = {c},
  column{5} = {c},
  column{6} = {c},
  column{9} = {c},
  column{10} = {c},
  cell{1}{1} = {r=2}{},
  cell{1}{2} = {r=2}{},
  cell{1}{3} = {r=2}{},
  cell{1}{4} = {c=4}{},
  cell{1}{8} = {c=4}{},
  cell{2}{4} = {c},
  cell{2}{8} = {c},
  cell{3}{1} = {r=12}{},
  cell{3}{2} = {c},
  cell{3}{3} = {r=2}{c},
  cell{3}{4} = {c},
  cell{3}{8} = {c},
  cell{4}{2} = {c},
  cell{4}{4} = {c},
  cell{4}{8} = {c},
  cell{5}{2} = {r=5}{c},
  cell{5}{3} = {c},
  cell{5}{4} = {c},
  cell{5}{8} = {c},
  cell{6}{3} = {c},
  cell{6}{4} = {c},
  cell{6}{8} = {c},
  cell{7}{3} = {c},
  cell{7}{4} = {c},
  cell{7}{8} = {c},
  cell{8}{3} = {c},
  cell{8}{4} = {c},
  cell{8}{8} = {c},
  cell{9}{3} = {c},
  cell{9}{4} = {c},
  cell{9}{8} = {c},
  cell{10}{2} = {r=5}{c},
  cell{10}{3} = {c},
  cell{10}{4} = {c},
  cell{10}{8} = {c},
  cell{11}{3} = {c},
  cell{11}{4} = {c},
  cell{11}{8} = {c},
  cell{12}{3} = {c},
  cell{12}{4} = {c},
  cell{12}{8} = {c},
  cell{13}{3} = {c},
  cell{13}{4} = {c},
  cell{13}{8} = {c},
  cell{14}{3} = {c},
  cell{14}{4} = {c},
  cell{14}{8} = {c},
  cell{15}{1} = {r=12}{},
  cell{15}{2} = {c},
  cell{15}{3} = {r=2}{c},
  cell{15}{4} = {c},
  cell{15}{8} = {c},
  cell{16}{2} = {c},
  cell{16}{4} = {c},
  cell{16}{8} = {c},
  cell{17}{2} = {r=5}{c},
  cell{17}{3} = {c},
  cell{17}{4} = {c},
  cell{17}{8} = {c},
  cell{18}{3} = {c},
  cell{18}{4} = {c},
  cell{18}{8} = {c},
  cell{19}{3} = {c},
  cell{19}{4} = {c},
  cell{19}{8} = {c},
  cell{20}{3} = {c},
  cell{20}{4} = {c},
  cell{20}{8} = {c},
  cell{21}{3} = {c},
  cell{21}{4} = {c},
  cell{21}{8} = {c},
  cell{22}{2} = {r=5}{c},
  cell{22}{3} = {c},
  cell{22}{4} = {c},
  cell{22}{8} = {c},
  cell{23}{3} = {c},
  cell{23}{4} = {c},
  cell{23}{8} = {c},
  cell{24}{3} = {c},
  cell{24}{4} = {c},
  cell{24}{8} = {c},
  cell{25}{3} = {c},
  cell{25}{4} = {c},
  cell{25}{8} = {c},
  cell{26}{3} = {c},
  cell{26}{4} = {c},
  cell{26}{8} = {c},
  hline{1,3,15,27} = {-}{},
  hline{2} = {4-11}{},
  hline{5,10,17,22} = {2-11}{},
}
\textbf{MT Direction} & \textbf{Models}        & \textbf{Methods}    & \textbf{Val 300} &                  &                  &                                & \textbf{FLORE 200} &                  &                  &                                                  \\
                      &                        &                     & \textbf{SPBLEU}  & \textbf{CharF++} & \textbf{Jaccard} & \textbf{\textbf{LLMaaJ}} & \textbf{SPBLEU}    & \textbf{CharF++} & \textbf{Jaccard} & \textbf{\textbf{\textbf{\textbf{LLMaaJ}}}} \\
EN-LB                 & Nllb-200-3.3B          & \textbf{BM}        & 19.97            & 37.03            & 0.27             &               0.75                 & 31.14              & 49.62            & 0.35             &         0.85                                         \\
                      & Llama-3.3-70B-Instruct &                     & 24.35            & 46.58            & 0.27             &              0.87                  & 22.55              & 43.08            & 0.26             &        0.83                                      \\
                      & Llama-3.2-3B-Instruct          & \textbf{BM} & 6.46             & 26.78            & 0.12             &               0.36                 & 4.80               & 22.10            & 0.09             &         0.36                                         \\
                      &                        & \textbf{DN}         & 37.98            & 55.41            & 0.37             &               0.82                 & 14.61              & 38.04            & 0.19             &        0.51                                          \\
                      &                        & \textbf{DL}         & 40.71            & 57.37            & 0.40             &              0.79                  & 20.93              & 41.51            & 0.22             &         0.52                                         \\
                      &                        & \textbf{DG}         & 42.01            & \textbf{57.89}   & 0.41             &                0.88              & 22.80              & 42.26            & 0.25             &            0.70                                   \\
                      &                        & \textbf{DGDC}       & \textbf{42.16}   & 57.87            & \textbf{0.42}    &              0.89                  & \textbf{23.40}     & \textbf{42.90}   & \textbf{0.26}    &          0.83                                      \\
                      &  Gemma-2-2b-it             & \textbf{BM} & 5.82             & 22.71            & 0.10             &                   0.50             & 4.61               & 20.78            & 0.07             &             0.51                                     \\
                      &                        & \textbf{DN}         & 41.77            & 57.71            & 0.42             &                0.89                & 20.41              & 41.21            & 0.25             &             0.78                                     \\
                      &                        & \textbf{DL}         & 43.78            & 59.02            & 0.44             &                0.87                & 24.03              & \textbf{42.95}   & \textbf{0.28}    &              0.79                                    \\
                      &                        & \textbf{DG}         & \textbf{44.58}   & \textbf{59.73}   & \textbf{0.45}    &                 0.87                & 23.47              & 42.72            & \textbf{0.28}    &              0.76                                    \\
                      &                        & \textbf{DGDC}       & 44.12            & 59.10            & \textbf{0.45}    &               0.90                 & \textbf{23.50}     & 42.49            & \textbf{0.28}    &             0.82                                     \\
LB-EN                 & Nllb-200-3.3B          & \textbf{BM}        &  40.51     & 56.81            & 0.48             &                     0.81           & 48.45              & 65.03            & 0.56             &                   0.85                               \\
                      & Llama-3.3-70B-Instruct &                     & 54.14            & 74.24            & 0.57             &               0.89                 & 33.96              & 58.02            & 0.41             &            0.86                                      \\
                      & Llama-3.2-3B-Instruct          & \textbf{BM} & 26.31            & 45.98            & 0.33             &              0.58                  & 17.62              & 36.79            & 0.26             &              0.46                                    \\
                      &                        & \textbf{DN}         & 42.78            & 59.33            & 0.48             &               0.82                 & 29.37              & 53.88            & 0.38             &               0.79                                   \\
                      &                        & \textbf{DL}         & 54.64            & 70.98            & 0.57             &                0.82                & 31.72              & 56.50            & 0.41             &              0.79                                    \\
                      &                        & \textbf{DG}         & \textbf{59.88}   & \textbf{74.97}   & \textbf{0.63}    &                 0.90               & \textbf{32.78}     & \textbf{57.69}   & \textbf{0.42}    &              0.81                                    \\
                      &                        & \textbf{DGDC}       & 57.88            & 73.46            & 0.60             &               0.89                 & 32.56              & 57.60            & 0.41             &              0.85                                    \\
                      & Gemma-2-2b-it            & \textbf{BM} & 27.11            & 47.44            & 0.34             &                   0.60             & 14.99              & 37.77            & 0.26             &      0.45                                           \\
                      &                        & \textbf{DN}         & 41.58            & 57.63            & 0.49             &              0.83                  & 42.46              & 60.55            & \textbf{0.51}    &         0.83                                         \\
                      &                        & \textbf{DL}         & 58.95            & 72.15            & 0.62             &                0.83                & 41.47              & 60.33            & 0.50             &           0.82                                       \\
                      &                        & \textbf{DG}         & \textbf{65.44}   & \textbf{76.96}   & \textbf{0.68}    &                 0.86               & 42.67              & \textbf{61.30}   & \textbf{0.51}    &           0.86                                       \\
                      &                        & \textbf{DGDC}       & 62.75            & 75.13            & 0.65             &               0.89                 & \textbf{42.73}     & 61.25            & \textbf{0.51}    &           0.85                                       
\end{tblr}}
\caption{This table presents the performance results obtained from training on datasets generated using different distillation models and methods. We report experimental results on two datasets, VAL 300 and FLORES 200. Additionally, we evaluated the performance of Nllb-200-3.3B and Llama-3.3-70B-Instruct on the same datasets, which strongly validate the effectiveness of our training approach. BM refers to the Base Model without any SFT. LLMaaJ refers to LLM-as-a-Judge, which gives a score from 0.0 to 1.0 with a granularity of 0.1.}
\label{tab:result_small}
\vspace{-0.3em}
\end{table*}
The latest open-source models are used as benchmark models, and their instruction-tuned versions are utilized to leverage their general capabilities in generating dialogues and answering questions. Based on the current leaderboard for Luxembourgish proficiency in LLMs~\cite{lothritz2025testing}, combined with the experimental results for the Germanic language group in Section~\ref{sec:lrlllm}, we select the top two base tiny models, which are Llama-3.2-3B-Instruct from Meta and Gemma-2-2b-it from Google.


The design of the input templates is considered crucial. In order to prevent the model from losing its general communication and generalization abilities after instruction tuning, it is necessary for prompts to be designed in alignment with chat templates that can be understood by the model. Based on this, basic prompt testing is conducted to identify the most suitable prompt for the model. Chat-based models have been observed to be prone to losing their communication capabilities after SFT, leading to the generation of endless content and a significant increase in the likelihood of hallucinations. Therefore, in the design of the questions, the corresponding starting prompts are set at the beginning of the model responses, such as "Here is the translation: {}". Through this linguistic guidance, the probability of hallucination is reduced and the model is also able to learn when to stop.

For the training data set, the LRL monolingual corpus is used primarily as the base material, from which the LRL-to-English mapping capability is distilled from larger models. As described in Section \ref{subsection: Data Collection}, publicly available press datasets and dictionary example sentences are utilized as the monolingual corpus, and distillation is performed using various teacher models. Finally, the correct word-to-word mapping capability is reinforced through the lemma search to verify the dictionary content. We classify fake targets distilled into four categories: fake targets obtained by distilling facebook/nllb-200-3.3B (\textbf{D}istill-\textbf{N}LLB, DN), the fake targets obtained by distilling meta-llama/Llama-3.3-70B-Instruct. (\textbf{D}istill-\textbf{L}lama, DL), the fake targets obtained by distilling GPT-4o-mini (\textbf{D}istill-\textbf{G}PT4O, DG), and the fake targets obtained after performing dictionary checking (\textbf{D}istill-\textbf{G}PT-\textbf{D}ict-\textbf{C}hecking, DGDC). Each category contains 621,033 data samples used for model training, all having the same LRL side texts, while the corresponding fake targets are generated by different teacher models.

For the validation set, the latest 300 press data entries (\textbf{Val 300}) from 2024 are used as monolingual corpus data, and the corresponding LRL entities are identified for the English mappings, thus preventing biases that may arise from the model having been trained on the validation dataset. And we also do a manual check for English translations. Furthermore, we utilize the FLORES-200 benchmark as an additional validation test set.

\subsection{Metrics}
There are multiple options of metrics available for MT tasks \cite{lo-etal-2023-beyond} and we mainly used the following three metrics for performance evaluation in our experiments: SPBLEU (SentencePiece BLEU), CharF++, and the Jaccard index. SPBLEU measures the similarity between machine translation outputs and reference translations using n-gram precision, employing a standardized SentencePiece model for subword tokenization and allowing effective differentiation between the performance of high-resource and low-resource languages, making it very valuable for comparative evaluation of multilingual models. CharF++ extends the character-level F score \cite{popovic-2015-chrf} metric used for machine translation evaluation, incorporating both character and word n-grams, showing a strong correlation with human judgments at both the system and the segment levels. The Jaccard index \cite{DBLP:journals/corr/abs-2110-09619} represents a fundamental statistical method to measure the similarity between sample sets, offering mathematical simplicity and interpretability, which makes it widely applicable across scientific disciplines. For LLMaaJ, we use google/gemma-3-27b-it as the judger throughout the entire paper.

\subsection{Results}
\begin{table*}[!h]

\begin{adjustbox}{width=\textwidth}
\renewcommand{\arraystretch}{0.9}
\begin{tabular}{ccccccccccc}
\toprule
\textbf{MT Direction }                      & \textbf{Model}            & \textbf{BOOLQ} & \textbf{CB}    & \textbf{COPA}  & \textbf{MULTIRC} & \textbf{RECORD} & \textbf{RTE}   & \textbf{WIC}   & \textbf{WSC}   & \textbf{AVG} \\
\midrule
\multirow{2}{*}{\textbf{BM}(Base Model)}   & Llama-3.2-3B-Instruct    & 0.62 & 0.55 & 0.71 & 0.52 & 0.41 & 0.64 & 0.51 & 0.28 & 0.53 \\
                                & Gemma-2-2b-it       & 0.73 & 0.55 & 0.86 & 0.81 & 0.56 & 0.82 & 0.49 & 0.56 & 0.67 \\
\midrule
\multirow{2}{*}{\textbf{En-LB}} & Llama-3.2-3B-Instruct-FT & 0.64 & 0.39 & 0.60 & 0.52 & 0.39 & 0.60 & 0.48 & 0.11 & 0.47 \\
                                & Gemma-2-2b-it-FT    & 0.71 & 0.52 & 0.89 & 0.75 & 0.41 & 0.72 & 0.51 & 0.49 & 0.62 \\
\midrule
\multirow{2}{*}{\textbf{LB-EN}} & Llama-3.2-3B-Instruct-FT & 0.64 & 0.30 & 0.69 & 0.51 & 0.46 & 0.62 & 0.52 & 0.24 & 0.50 \\
                                & Gemma-2-2b-it-FT    & 0.69 & 0.25 & 0.90 & 0.76 & 0.45 & 0.73 & 0.51 & 0.43 & 0.59 \\
\bottomrule
\end{tabular}
\end{adjustbox}
\caption{Variations in overall performance on the SuperGLUE benchmark before and after distillation training, evaluating whether fine-tuning on LRLs induces catastrophic forgetting. The model names appended with the suffix ``-FT'' denote the models after applying the proposed distillation fine-tuning method.}
\label{tab:forgotting}
\end{table*}

\subsubsection{Performance gain}

    The results in Table \ref{tab:result_small} clearly demonstrate that fine-tuning in both translation directions is highly effective. For example, the baseline EN$\to$LB models exhibit SPBLEU scores around 30, but after fine-tuning, these scores increase to nearly 38--40 values approaching our threshold for high-quality translations (SPBLEU $>$ 40). In contrast, LB$\to$EN translations consistently score above 40, yet generating fluent Luxembourgish in the EN$\to$LB direction remains a significant challenge. Furthermore, our experiments indicate that even a 3B model, when effectively distilled, can rival or even surpass larger models in low-resource language translation tasks. Our results indicate that GPT-4o-based distillation methods, in particular, produce substantial improvements in translation quality, confirming that parallel corpora generated by LLM represent a viable and promising strategy for supporting LRL translation tasks. In order to validate the model translation performance, we also extracted a portion of the data and asked Luxembourgers who are at least bilingual in Luxembourgish and English to label it as ground truth as data quality validation. The SPBLEU score achieved with these labeled data was 51.08 on our fine-tuned Gemma~2--2b, showing a comparable score calculated using GPT-generated data as ground truth. Regarding the LLMaaJ score of the model, we obtained performance evaluation results and trends that are largely consistent with those of the SPBLEU parameter, further cross-validating the feasibility of LLMaaJ. However, since LLMs are black-box models with limited interpretability, the scores produced by LLMaaJ can only serve as a reference and do not guarantee accuracy or validity.

    \textbf{To address RQ2}, the performance improvement of the model after fine-tuning with data distillation enhancement is highly significant. For the two tested models, the performance gains are reflected in SPBLEU scores that surpass those of certain expert translation models. Furthermore, the enhancement observed in the EN$\to$LB direction is greater than that in the reverse direction, further strengthening the model’s ability to generate Luxembourgish. Therefore, LRLs can substantially improve the translation capacity of the model for low-resource languages, and even smaller models can achieve promising results.

\subsubsection{Does data size really matter?}
\begin{figure}[!h]
    \hspace*{-0.4cm}
    \includegraphics[width=1.10\linewidth]{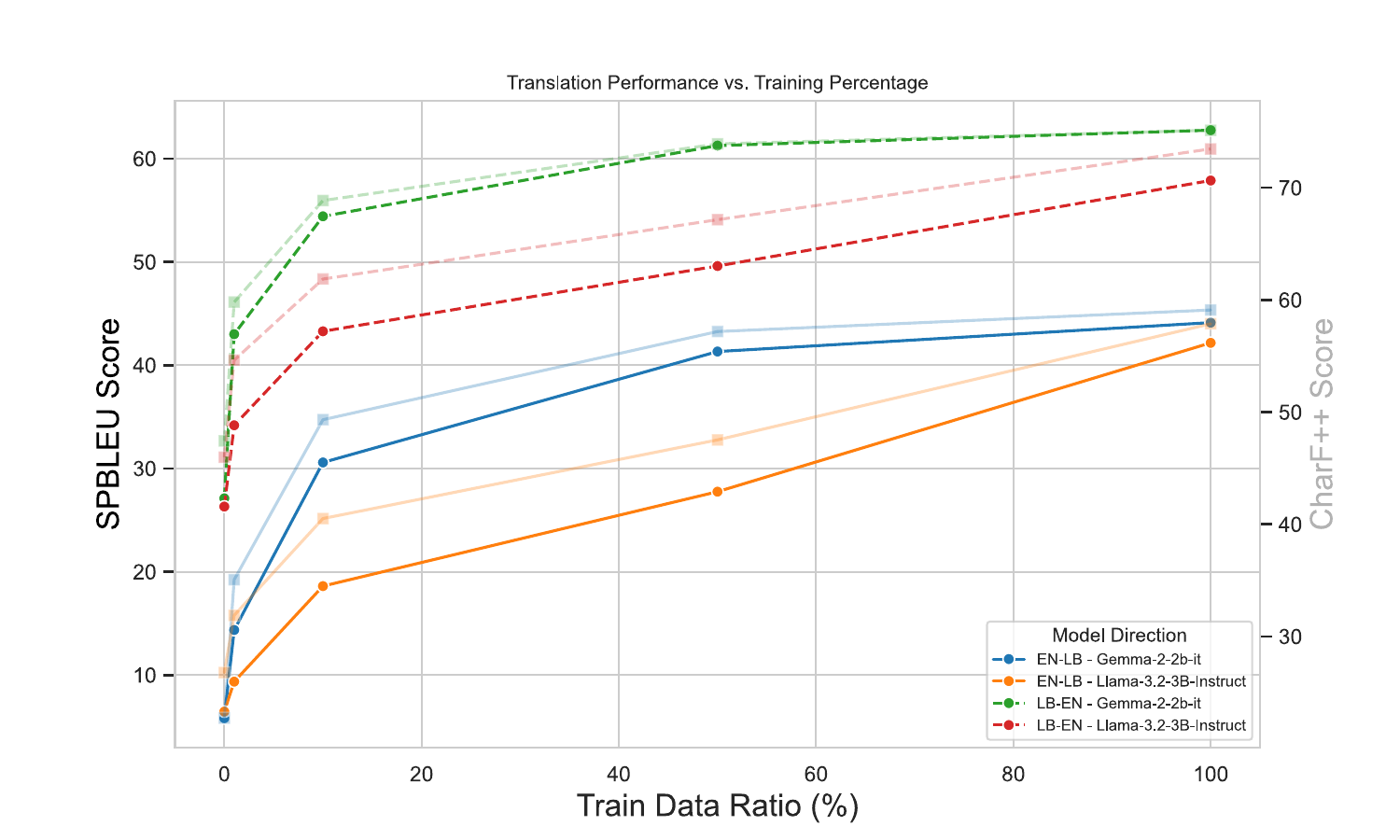}
    \caption{Performance as a function of training data size ratio. The dashed lines with transparency indicate the trend of CharF++ scores, while the solid lines represent the SPBLEU scores. The x-axis denotes the proportion of the training data relative to the full original training set.}
    \label{fig:datasize}
\end{figure}

Figure~\ref{fig:datasize} illustrates the strong influence of the size of the data set on the quality of the translation in both directions (English$\Leftrightarrow$Luxembourgish), more detailed data in the Appendix Table \ref{tab:size}. Even using as little as 1\% of the available data yields modest improvements over the base model, yet the most substantial gains emerge only at higher data ratios. For example, increasing the data from 25\% to 100\% nearly doubles SPBLEU in the EN$\rightarrow$LB direction for both Llama~3.2--3B and Gemma~2--3B. Notably, Gemma~2--3B seems to learn faster in the lower data regimes, but shows some performance attenuation beyond the 50\% threshold.


\subsubsection{Catastrophic forgetting?}

As a general-purpose model, it is capable of not only performing translation tasks but also handling multiple tasks such as planning, solving mathematical problems, coding, etc., other than translation. However, after training the model specifically for translation purposes, a critical question arises: Does the model suffer catastrophic forgetting? This issue is of urgent concern and has significant implications for the potential of the model for generalized usage. To investigate this, we compared the model performance with the SuperGLUE benchmark \cite{sarlin2020superglue} before and after training which is a widely adopted benchmark suite for evaluating LLM general performance. Table \ref{tab:forgotting} presents the performance results, indicating that fine-tuning, while enhancing translation capabilities, has a minimal impact on the model's proficiency in other tasks, demonstrating its robustness and adaptability. The analysis confirms that distillation can enhance translation performance while preserving the overall aptitude of the model across various tasks

\subsubsection{Can we do LoRA?}

We also carried out experiments using the same data to assess how the LoRA (Low-Rank Adaptation \cite{hu2022lora}) rank parameter influences training performance in translation tasks involving Luxembourgish and English. Specifically, we evaluated the ranks 8, 16, 32 and 64 in our models. The results, presented in Table \ref{tab:lora-ft-spbleu} and \ref{tab:lora-ft}, indicate that variations in the LoRA rank parameter have a minimal influence on the overall translation performance, with differences typically within 1 to 2 SPBLEU points. More importantly, models fine-tuned using LoRA consistently underperformed compared to their fully fine-tuned counterparts, achieving notably lower BLEU scores compared to table \ref{tab:result_small}. Moreover, after LoRA-based SFT, we also observed an increased tendency toward hallucination. Due to the consistently lower performance and negligible differences observed among the varying LoRA ranks, we opted not to use LoRA fine-tuning in machine translation tasks. Instead, we focused on full-model fine-tuning, which demonstrated significantly better results. These findings suggest that, while LoRA provides computational efficiency, its limited parameter updates are insufficient to capture the nuanced linguistic features required for effective translation of LRLs and may even be harmful

\begin{table}[htbp]
\centering
\caption{Impact of LoRA Rank on SPBLEU During Fine-Tuning, Evaluated Across Three Rank Values}
\label{tab:lora-ft-spbleu}
\resizebox{1.0\columnwidth}{!}{%
\begin{tabular}{ccc c}
\toprule
\multirow{2}{*}{\textbf{EN-LB}} &
\multirow{2}{*}{\textbf{Rank (LoRA)}} &
\textbf{Val 300} &
\textbf{FLORE 200} \\
& & \textbf{SPBLEU} & \textbf{SPBLEU} \\
\midrule
\multirow{4}{*}{Llama 3.2-3B} 
 & \textbf{Base Model} & 6.46  & 4.80 \\
 & \textbf{32}     & 12.95 & 9.46 \\
 & \textbf{64}     & 13.05 & 9.23 \\
 & \textbf{128}    & 13.32 & 9.27 \\
\midrule
\multirow{4}{*}{Gemma 2-2B}

 & \textbf{Base Model} & 5.82  & 4.61 \\
 & \textbf{32}     & 13.07 & 8.88 \\
 & \textbf{64}     & 13.17 & 9.12 \\
 & \textbf{128}    & 13.31 & 9.21 \\
\bottomrule
\end{tabular}
}
\end{table}

\subsubsection{How about other LRLs?}

We demonstrate that distillation from various large teacher models can elevate the low-resource translation performance of smaller models to a level comparable to that of expert systems, thereby confirming the potential of small models in translation tasks. To further verify the generality of our findings, we additionally extracted 10{,}000 sentences from the WMT~2025 in Khasi, Assamese, and Ukrainian (Facebook-WikiMatrix-1-eng-ukr subset filtered for sentence lengths between 200 and 299 tokens), along with 1{,}000 pairs of corresponding sentences as a validation set. 

Using the same methodology, we performed data distillation for one-sided sentences with three different models: the previously mentioned NLLB model, the Llama 3.3-70B model, and GPT-4o-mini. We then trained Llama-3.2-3B with identical prompts and evaluated performance on the validation set using the corresponding ground-truth annotations provided by the dataset.

\begin{figure*}[!htbp]
    \hspace*{-0.4cm}
    \includegraphics[width=1.00\linewidth]{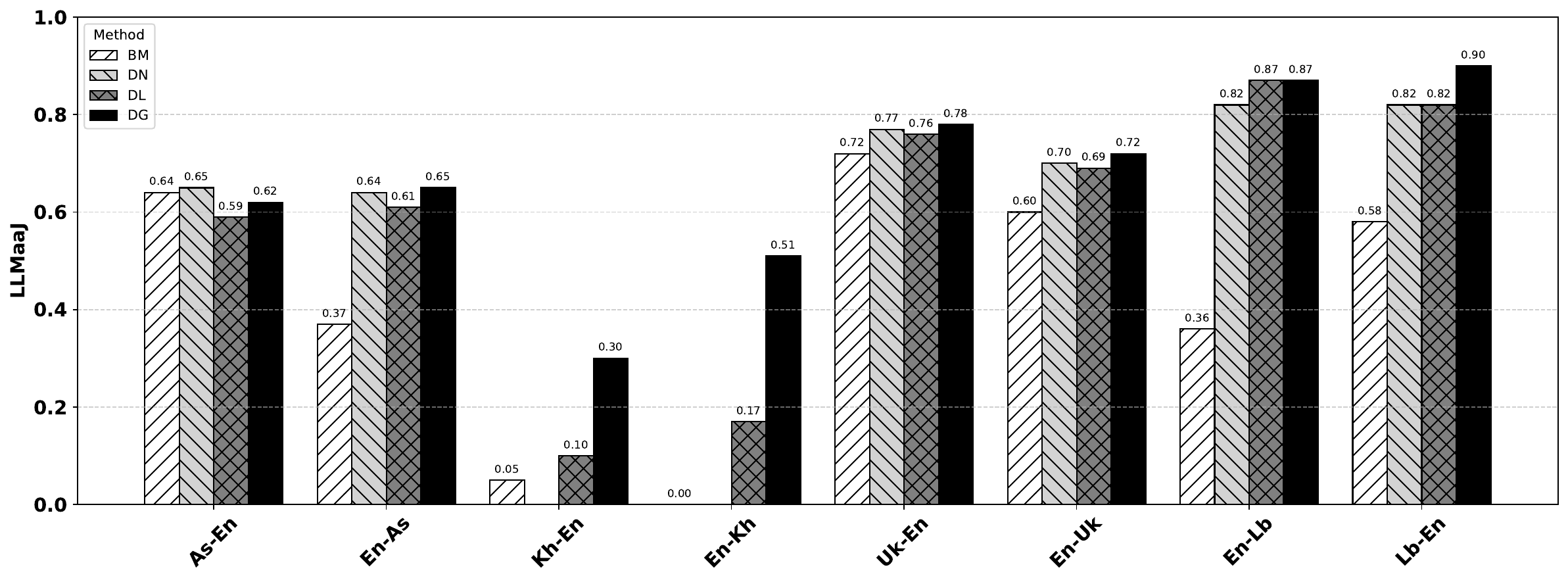}
    \caption{This figure compares the performance of four LRL pairs under the base model (Llama-3.2-3B) and under knowledge distillation from different teacher models, evaluated using the LLMaaJ metric. ``As'' denotes Assamese, ``Kh'' denotes Khasi, and ``Uk'' denotes Ukrainian. Notably, the Kh–-En and En-–Kh directions lack results for the DN setting (i.e., using NLLB-200-3.3B as the teacher model), as NLLB does not provide support for Khasi.}
    \label{fig:LRL_general}
\end{figure*}

As shown in Figure~\ref{fig:LRL_general} and Table~\ref{tab:general_langauge_results}, when the model performance is already high---such as in the As--En direction, where the base model reaches a score of 0.64---the effect of distillation is not pronounced. In contrast, for the En--As, En--Kh, En--Lb, and Lb--En directions, the results reveal that distillation from the teacher model is critical, leading to substantial improvements in translation quality. This suggests that distilled data can effectively impart knowledge of resource-scarce languages to small models, with minimal degradation in their general performance.

\section{Conclusion}


In this study, drawing on evaluations conducted with the FLORES-200 dataset within the scope of \textbf{RQ1}, we quantitatively assessed both open source and closed source state-of-the-art models and examined the performance distribution of LLM support for different language families. Our findings provide quantitative evidence of the current state of technological inequity. Unfortunately, LLMs are far from a universal solution for certain endangered LRLs at present.

To address \textbf{RQ2}, we used Luxembourgish as a representative case and primarily leveraged a monolingual Luxembourgish corpus---combined with knowledge distillation and data augmentation techniques---to fine-tune compact 2B/3B-scale LLMs for bidirectional translation between Luxembourgish and English. Our findings reveal that, although smaller models inherently face greater challenges in processing LRLs compared to larger architectures, targeted distillation markedly enhances their performance. The fine-tuned models deliver usable and reliable translations, validated not only through conventional semantic evaluation metrics but also via \textit{LLM-as-a-Judge}, an LLM-based evaluation framework.

Finally, to provide further insights into distillation within \textbf{RQ3}, our analysis highlights the influence of the size of the dataset, revealing that even minimal data augmentation can substantially improve translation performance, while marginal gains from additional data tend to diminish as the dataset grows. We also verified that this approach generalizes to various low-resource languages, with only a minor impact on their overall capabilities after training. Lastly, in the context of translation tasks, the use of LoRA is not recommended, as it yields limited performance improvements.

In general, the rapid advancement of LLMs has not been fully extended to LRLs and remains geographically correlated with regional development levels. Although linguistic proximity to HRL can partially alleviate this gap, the majority of LRLs are still not supported by current LLMs. This underscores both the inherent limitations of LLMs and the urgency of promoting greater technological equity. Moreover, the findings of this study offer practical guidance for developing portable and cost-efficient translation models that effectively support selected LRLs while preserving the overall capabilities of the models, thereby pointing toward a promising direction for advancing LRL translation through transformer-based approaches.

\section*{Limitations}

This study has several limitations that should be considered. Firstly, despite efforts to gather diverse data sources, the dataset size and diversity for Luxembourgish remain constrained compared to high-resource languages. As a result, the generalizability of our findings might be limited. Additionally, our reliance on knowledge distillation from large pre-trained models assumes access to high-quality pretrained models, which may not be feasible in all low-resource contexts. Lastly, translation performance metrics such as BLEU scores may not fully capture the nuanced linguistic accuracy or cultural appropriateness of translations, necessitating complementary qualitative assessments in future studies. Moreover, future work may explore the validation of fine-tuning and distillation performance on more LRLs, as well as on artificially constructed languages, such as Elvish. For LRL analysis, we still have a lot of LRLs that have not been considered, like the Khoe languages branch in Namibia.

\section*{Ethics Statement}

All models and resources developed in this work are strictly intended for research and educational purposes according to OpenAI usage guidelines; no model weights or derivatives are used — or will be used — for any commercial application. We exclusively utilize publicly available corpora or datasets for which explicit authorization has been obtained from the original data providers.  All license terms have been reviewed to ensure full compliance with copyright, attribution, and sharing requirements.

No personally identifiable information (PII) is collected during this research. All data processing, storage, and retention policies are fully aligned with the EU General Data Protection Regulation (GDPR). The dataset of LOD.lu is under the CC0 license. As most of RTL datasets are based on articles from RTL, we cannot publish them, but we make them available to researchers on request.

All code, models, and processed data artifacts will be released under an open-source, research-oriented license (e.g., CC BY-NC), accompanied by comprehensive documentation and bias-analysis methodology to promote transparency and reproducibility.  We commit to ongoing ethical oversight through periodic reevaluation of datasets and model outputs, prompt updates in response to emerging concerns, and consultation with interdisciplinary advisory boards to ensure adherence to the highest ethical standards.

\bibliography{custom}

\appendix

\label{sec:appendix}
\section*{Appendix}
\section{Data Processing}

Dataset selection directly impacts the reliability and generalizability of experimental results. Our criteria include having enough test samples, providing reference responses, and minimizing potential biases from overlap with pre-training data.

FLORES-200 \cite{nllb2022no} is a benchmark dataset specifically designed for low-resource and multilingual machine translation, serving as an extended version of FLORES-101 \cite{DBLP:journals/corr/abs-2106-03193}. It covers 200 languages and consists of sentences extracted from 842 web articles, with an average length of approximately 21 words. These sentences are divided into three datasets: dev, devtest, and a hidden test set. Since we require additional evaluation metrics, we use devtest as our set of tests in this study. In our paper, we primarily evaluate the translation performance of all 200 languages into English. However, in the subsequent model training, we focus solely on the Luxembourgish-English language pair for training and testing.

The VAL 300 validation set was constructed using 300 pieces of official news content from July 2024 as the source data. The corresponding ground truth in Luxembourg was generated using ChatGPT, followed by dictionary-based verification to ensure validity. Furthermore, we extracted 30 samples from the dataset and engaged Luxembourgish-English bilingual speakers to perform a quality assessment.



\section{Experiments settings}
\label{apd:expset} 

In our experiments, we used primarily two distinct models for supervised fine-tuning (SFT) to evaluate performance and optimization strategies. To ensure an effective training process, several hyperparameters and model configurations were meticulously selected. Specifically, the warm-up ratio was set to 0.5, facilitating a gradual increase in the learning rate during the initial training phase for improved convergence stability. The maximum gradient norm was restricted to 0.3, serving as a mechanism to prevent excessively large parameter updates and promote stable optimization dynamics. Furthermore, the input sequence length was capped at 512 tokens, ensuring that all processed data adhered to this fixed-length constraint. A weight decay of 0.01 was applied to regularize the model parameters and mitigate the risk of overfitting. It is worth noting that all of our models were trained for only one epoch. This decision was based on our observation that evaluation metrics reached their optimal performance after a single epoch, while additional epochs exacerbated the impact of noisy data without yielding performance improvements.

To ensure reproducibility across experiments, a fixed random seed of 3407 was utilized. For model architecture selection, two distinct approaches were considered: standard fine-tuning and LoRA. In cases where LoRA was employed, specific layers were targeted for adaptation, including "q\_proj," "k\_proj," "v\_proj," "o\_proj," "gate\_proj," "up\_proj," and "down\_proj." The LoRA alpha parameter was configured to a value of 8, while the dropout rate for LoRA layers was set to 0, indicating that no dropout-based regularization was applied to these low-rank adaptation layers.

For tokenization and input preparation, a standardized procedure was adopted to ensure consistency in sequence length across the examples. The tokenizer processed each input field by truncating sequences exceeding the maximum length of 512 tokens and padding shorter sequences to this fixed length. This was achieved using the `padding="max\_length"` option, thereby guaranteeing uniformity in input representation prior to model training. During the inference stage, we set the temperature parameter to 0.1 (close to 0), which has been shown to help achieve optimal machine translation performance~\cite{LI2025242}. In addition, we set \texttt{max\_new\_tokens} to 512, enable \texttt{do\_sample = True}, and set \texttt{top\_p = 0.9}.

\begin{table}[htbp]
\centering
\adjustbox{max width=\linewidth}{
\begin{tabular}{lll}
\toprule
\textbf{Model}        & \textbf{Reference}        & \textbf{SFT Methods} \\ 
\midrule
Llama-3.2-3B-Instruct & \citealp{Llama-3.2-3B-Instruct} & FS/ LoRA SFT  \\  
gemma-2-2b-it         & \citealp{gemma-2-2b-it} & FS/ LoRA SFT  \\  
\bottomrule
\end{tabular}
}
\caption{Various models and their SFT methods. "FS/ Lora SFT" refers to full-size and "Lora SFT" denotes Low-Rank Adaptation SFT only.}
\label{tab: models}
\vspace{-1em}
\end{table}

\section{Dictionary Processing}
\begin{figure*}[!htbp]
    \vspace{-1em}
    \centering
    \includegraphics[width=\linewidth]{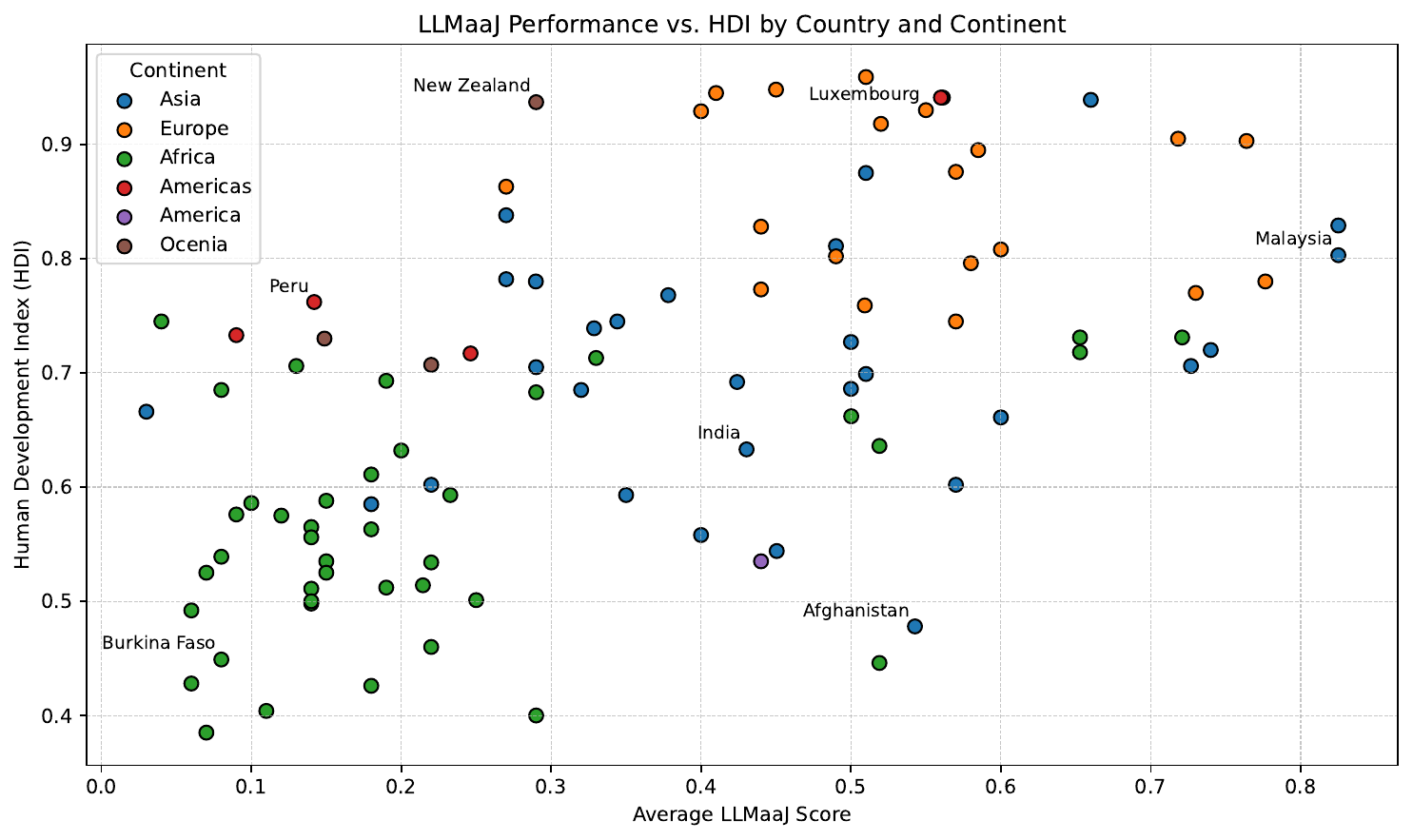}
    \caption{Scatter Plot of LLMaaJ Score and HDI Relation for LRLs}
    \label{fig:scatter}
    \vspace{-1em}
\end{figure*}
\label{apd:dictrag}

In our approach to enhancing translation accuracy, particularly for Luxembourgish, we developed a retrieval pipeline using Haystack 2.0. The pipeline utilizes a BM25 retriever to identify relevant dictionary entries that align closely with the input text. The retrieved dictionary entries are then incorporated directly into the prompt provided to GPT-4O, offering multiple lexical choices that help clarify ambiguous terms.

This method operates as follows: first, the BM25 retriever ranks and returns the most relevant dictionary entries based on the Luxembourgish input. These entries serve as additional context within the prompt, guiding GPT-4o toward more accurate translations. Subsequently, the original Luxembourgish sentence and the relevant dictionary context are submitted to GPT-4o for translation. By explicitly integrating these dictionary options into the prompt, GPT-4o is better equipped to resolve lexical ambiguities and correct potential translation errors, enhancing translation accuracy and coherence.

\section{Language Ability On LLMs}
\subsection{Translation Performance and Human Development Disparities}
\label{apd:ability}
In this analysis, LRLs are operationally defined as those that comprise less than 0.1\% of web content (according to W3Techs statistics\footnote{\url{https://w3techs.com/technologies/overview/content_language}}). The average \textit{LLMaaJ} scores were calculated exclusively for the selected LRLs that also exist in the FLORES-200 dataset. Country - LRLs pairs were identified based on a mapping that utilizes Wikipedia-derived estimates of language speaker distribution. 

Figure \ref{fig:scatter} reveals a clear positive correlation between a country’s human development level (HDI) and the translation quality of its low-resource languages as judged by LLMs. Each point in the scatter represents a FLORES-200 language linked to a country’s HDI, and the overall trend slopes upward – higher-HDI countries tend to have languages with higher LLMaaJ translation scores. This suggests that socioeconomic factors underpin disparities in LLM translation coverage, echoing the “digital language divide” observed in AI research~\cite{okolo2024closing}. In other words, languages from more developed regions generally receive far better support in large multilingual models than those from less developed regions. 

When grouping languages by development tiers, the performance gap is stark. Languages from Very High HDI countries (HDI $\geq$ 0.80) achieve an average LLMaaJ score of around 0.54, more than double the ~0.22 average for languages from Low HDI countries (HDI < 0.55). Median scores likewise jump from only ~0.15 in low-HDI settings to ~0.53 in very-high-HDI settings. This means a typical low-resource language in a highly developed society enjoys significantly better machine translation quality than one in a low-development context. Crucially, it is not simply the number of speakers but the socioeconomic context and digital resources that dictate how well a language is served by AI. For instance, Hindi (with over 500 million speakers) has historically been treated as “low-resource” for NLP, whereas a smaller language like Dutch (with a fraction of the speakers, but backed by a high-HDI country) is well-supported. The greater availability of data and funding in high-HDI environments allows LLMs to achieve markedly better translations for those languages. 

Geographic disparities are especially pronounced. Nearly all African languages in the study cluster toward the lower-left of Figure \ref{fig:scatter}, indicating both low HDI and poor translation performance. In fact, none of the African languages evaluated approach the top tier of LLMaaJ scores – a finding consistent with reports that even state-of-the-art multilingual models still lag on African languages due to limited training data and quality. By contrast, European languages (from countries with generally high HDI) occupy the upper range of the plot; these languages achieve some of the highest scores (e.g. minority languages like Occitan in France reach LLMaaJ $\approx0.76$). Several Asian languages spoken in high-HDI regions likewise perform strongly – for example, Standard Malay (Malaysia/Brunei) attains average scores above 0.80 in our data. Meanwhile, many languages of low-HDI countries remain at the bottom: Dzongkha of Bhutan (medium HDI) has one of the lowest scores (LLMaaJ $\approx0.03$), and numerous Sub-Saharan African languages (e.g. Tigrinya of Eritrea) register below 0.10. These patterns suggest that languages benefiting from a robust digital infrastructure or from close linguistic ties to well-resourced tongues (as Occitan does to French) see far better outcomes, whereas languages in impoverished or isolated settings are left behind. 

Overall, the strong HDI-performance correlation highlights a systemic inequality in LLM coverage. The correlation coefficient score between HDI and LLMaaJ average score is 0.566, indicating a medium-high correlation. Communities in low-development regions face a double disadvantage: they are underserved by technology on top of existing socio-economic challenges. Indeed, globally fewer than 1\% of languages have sufficient data to be considered high-resource, leaving speakers of the other 99\% “essentially cut off from global technological progress”. This lack of access to quality translation and language tools can hinder information access, education, and opportunities, thereby exacerbating the digital divide and reinforcing global inequalities. Our findings underscore that current multilingual AI models, despite their broad reach, de facto offer far stronger support for languages of wealthy, high-HDI communities than for those of poorer regions. Addressing this gap will require concerted efforts to bring truly inclusive language coverage to the forefront, rather than merely adding more languages without improving quality for the most disadvantaged.

\subsection{Result Tables}
\label{tab:llmasjudge}

\onecolumn
{\tiny
\centering
\setlength\tabcolsep{3pt}

\begin{longtable}{|l|l|l|l|l|l|l|l|l|l|l|l|l|}

\caption{The LLMaaJ results on the FLORES-200 dataset are derived from evaluations of 10 distinct large language models. Population estimates are based on heterogeneous sources, and the reported population are not guaranteed to be accurate. Therefore, they should be interpreted with appropriate caution. }\label{tab:LLMAAJALL}\\

\hline
Language Name &
  \begin{tabular}[c]{@{}l@{}}Language\\ Branch\end{tabular} &
  Population &
  \begin{tabular}[c]{@{}l@{}}GPT4o\\ mini\end{tabular} &
  \begin{tabular}[c]{@{}l@{}}Llama-\\ 3.1-8B\end{tabular} &
  \begin{tabular}[c]{@{}l@{}}Llama-\\ 3.2-3B\end{tabular} &
  \begin{tabular}[c]{@{}l@{}}Ministral\\ -8B\end{tabular} &
  Phi-3 &
  Phi-3.5 &
  \begin{tabular}[c]{@{}l@{}}Qwen2.5\\ -1.5B\end{tabular} &
  \begin{tabular}[c]{@{}l@{}}Qwen2.5\\ -3B\end{tabular} &
  \begin{tabular}[c]{@{}l@{}}gemma2\\ -2b\end{tabular} &
  \begin{tabular}[c]{@{}l@{}}gemma2\\ -9b\end{tabular} \\ \hline
Central Atlas Tamazight        & \multirow{4}{*}{Berber}        & ~3-4 million        & 0.017 & 0.008 & 0.006 & 0.008 & 0.007 & 0.014 & 0.006 & 0.01  & 0.011 & 0.014 \\ \cline{1-1} \cline{3-13} 
Kabyle                         &                                & ~5 million          & 0.078 & 0.054 & 0.027 & 0.025 & 0.02  & 0.038 & 0.02  & 0.042 & 0.028 & 0.08  \\ \cline{1-1} \cline{3-13} 
Tamasheq (Latin script)        &                                & ~500,000            & 0.143 & 0.101 & 0.067 & 0.082 & 0.088 & 0.093 & 0.061 & 0.09  & 0.096 & 0.142 \\ \cline{1-1} \cline{3-13} 
Tamasheq (Tifinagh script)     &                                & ~500,000            & 0.021 & 0.009 & 0.007 & 0.009 & 0.008 & 0.022 & 0.005 & 0.013 & 0.016 & 0.018 \\ \hline
Hausa                          & Chadic                         & ~40 million         & 0.774 & 0.534 & 0.166 & 0.132 & 0.089 & 0.101 & 0.082 & 0.11  & 0.228 & 0.656 \\
\hline

\end{longtable}
\label{tab:results}

\begin{longtable}{|l|l|l|l|l|l|l|l|l|l|l|l|l|}

\caption{The Corpus BLEU results on the FLORES-200 dataset are derived from evaluations of 10 distinct large language models. Population estimates are based on heterogeneous sources, and the reported population are not guaranteed to be accurate. Therefore, they should be interpreted with appropriate caution. }
\label{tab:BLEU}\\
\hline
    Language Name &
  \begin{tabular}[c]{@{}l@{}}Language \\ Branch\end{tabular} &
  Population &
  \begin{tabular}[c]{@{}l@{}}GPT4o\\ Mini\end{tabular} &
  \begin{tabular}[c]{@{}l@{}}Llama\\ 3.1 8B\end{tabular} &
  \begin{tabular}[c]{@{}l@{}}Llama\\3.2 3B\end{tabular} &
  \begin{tabular}[c]{@{}l@{}}Ministral\\ 8B\end{tabular} &
  Phi-3 &
  Phi-3.5 &
  \begin{tabular}[c]{@{}l@{}}Qwen2.5\\ 1.5B\end{tabular} &
  \begin{tabular}[c]{@{}l@{}}Qwen2.5\\ 3B\end{tabular} &
  \begin{tabular}[c]{@{}l@{}}gemma-2\\ 2B\end{tabular} &
  \begin{tabular}[c]{@{}l@{}}gemma-2\\ 9B\end{tabular} \\ \hline
Central Atlas Tamazight        & \multirow{4}{*}{\textbf{Berber}}             & ~3-4 million          & 1.4  & 0.4  & 0.4  & 0.2  & 1.0  & 0.8  & 0.2  & 0.8  & 0.4  & 1.4  \\ \cline{1-1} \cline{3-13} 
Kabyle                         &                                     & ~5 million            & 4.0  & 3.3  & 1.4  & 0.9  & 1.7  & 0.7  & 0.5  & 1.5  & 1.4  & 4.3  \\ \cline{1-1} \cline{3-13} 
Tamasheq (Latin script)        &                                     & ~500,000              & 5.2  & 3.9  & 2.7  & 1.9  & 4.3  & 1.7  & 1.0  & 3.4  & 3.3  & 4.9  \\ \cline{1-1} \cline{3-13} 
Tamasheq (Tifinagh script)     &                                     & ~500,000              & 1.3  & 0.4  & 0.3  & 0.2  & 1.0  & 0.7  & 0.1  & 0.5  & 0.6  & 1.1  \\ \hline
Hausa                          & Chadic                              & ~40 million           & 30.4 & 20.0 & 7.5  & 2.9  & 3.9  & 1.6  & 1.5  & 4.5  & 8.9  & 25.9 \\
\hline
\end{longtable}%
\twocolumn
}


\begin{table*}[htbp]
\centering
\caption{Impact of LoRA Rank on Performance During Fine-Tuning, Evaluated Across Three Rank Values}
\label{tab:lora-ft}
\resizebox{2\columnwidth}{!}{%
\begin{tabular}{ccccc cccc}
\toprule
\multirow{2}{*}{\textbf{EN-LB}} &
\multirow{2}{*}{\textbf{Rank (LoRA)}} &
\multicolumn{3}{c}{\textbf{Val 300}} &
\multicolumn{3}{c}{\textbf{FLORE 200}} \\
\cmidrule(lr){3-5} \cmidrule(lr){6-8}
 & & \textbf{SPBLEU} & \textbf{CharF++} & \textbf{Jaccard} 
   & \textbf{SPBLEU} & \textbf{CharF++} & \textbf{Jaccard} \\
\midrule
\multirow{4}{*}{Llama 3.2-3B} 
 & \textbf{Base Model} & 6.46 & 26.78 & 0.12 & 4.80 & 22.10 & 0.09 \\
 & \textbf{r = 32}     & 12.95 & 33.09 & 0.19 & 9.46 & 29.64 & 0.14 \\
 & \textbf{r = 64}     & 13.05 & 33.59 & 0.19 & 9.23 & 28.93 & 0.14 \\
 & \textbf{r = 128}    & 13.32 & 34.09 & 0.20 & 9.27 & 29.16 & 0.14 \\
\midrule
\multirow{4}{*}{Gemma 2-2B} 
 & \textbf{Base Model} & 5.82 & 22.71 & 0.10 & 4.61 & 20.78 & 0.07 \\
 & \textbf{r = 32}     & 13.07 & 33.36 & 0.21 & 8.88 & 27.93 & \textbf{0.16} \\
 & \textbf{r = 64}     & 13.17 & 33.35 & 0.21 & 9.12 & 28.06 & 0.16 \\
 & \textbf{r = 128}    & 13.31 & \textbf{33.69} & 0.21 & 9.21 & 28.20 & 0.16 \\
\bottomrule
\end{tabular}
}
\end{table*}

\section{Dataset Size Influence}
Table~\ref{tab:size} in the appendix presents a comprehensive analysis of how dataset size influences translation performance in our low-resource Luxembourgish-English setting. We experimented with dataset sizes ranging from as small as 1\% to the full dataset (100\%). The results demonstrate a clear, positive correlation between the amount of data utilized during fine-tuning and the subsequent translation quality, as measured by BLEU scores.

In both translation directions (EN$\rightarrow$LB and LB$\rightarrow$EN), we observed that even very small datasets (e.g., 1\%–5\%) provide measurable improvements over baseline models, indicating that the models begin acquiring beneficial linguistic patterns early in the fine-tuning process. However, substantial performance gains occur predominantly when increasing the dataset size beyond 25\%. For instance, moving from 25\% to 100\% dataset size nearly doubles the SPBLEU scores for the EN$\rightarrow$LB direction, clearly highlighting the significance of sufficient data availability for generating fluent, accurate translations in low-resource languages.

Interestingly, the Gemma-2-2B model displayed a relatively faster learning trajectory compared to the Llama-3.2-3B model in smaller data regimes (below 50\%). Nevertheless, Gemma-2-2B exhibited a notable attenuation in performance improvements beyond the 50\% data threshold, suggesting a diminishing return effect when datasets grow larger. Conversely, the Llama-3.2-3B model showed steadier improvements without significant attenuation up to the full dataset size, potentially indicating better scalability of linguistic capabilities with increased training data.

\begin{table*}[htbp]
\caption{Impact of Dataset Size on the Performance of Fine-Tuning}
\label{tab:size}
\resizebox{2\columnwidth}{!}{%
\begin{tabular}{|c|c|ccc|ccc|}
\hline
\multirow{2}{*}{\textbf{\begin{tabular}[c]{@{}c@{}}English to\\  Luxembourgish\end{tabular}}} &
  \multirow{2}{*}{\textbf{\begin{tabular}[c]{@{}c@{}}Dataset\\ Ratio\end{tabular}}} &
  \multicolumn{3}{c|}{\textbf{Val 300}} &
  \multicolumn{3}{c|}{\textbf{FLORE 200}} \\ \cline{3-8} 
 &
   &
  \multicolumn{1}{c|}{\textbf{SPBLEU}} &
  \multicolumn{1}{c|}{\textbf{CharF++}} &
  \textbf{Jaccard} &
  \multicolumn{1}{c|}{\textbf{SPBLEU}} &
  \multicolumn{1}{c|}{\textbf{CharF++}} &
  \textbf{Jaccard} \\ \hline
\multirow{5}{*}{Llama 3.2 -3B} &
  \textbf{0\%} &
  \multicolumn{1}{c|}{6.46} &
  \multicolumn{1}{c|}{26.78} &
  0.12 &
  \multicolumn{1}{c|}{4.80} &
  \multicolumn{1}{c|}{22.10} &
  0.09 \\ \cline{2-8} 
 &
  \textbf{1\%} &
  \multicolumn{1}{c|}{9.36} &
  \multicolumn{1}{c|}{31.88} &
  0.16 &
  \multicolumn{1}{c|}{6.53} &
  \multicolumn{1}{c|}{26.31} &
  0.10 \\ \cline{2-8} 
 &
  \textbf{10\%} &
  \multicolumn{1}{c|}{18.61} &
  \multicolumn{1}{c|}{40.51} &
  0.23 &
  \multicolumn{1}{c|}{9.79} &
  \multicolumn{1}{c|}{30.65} &
  0.14 \\ \cline{2-8} 
 &
  \textbf{50\%} &
  \multicolumn{1}{c|}{\textbf{27.75}} &
  \multicolumn{1}{c|}{\textbf{47.52}} &
  \textbf{0.30} &
  \multicolumn{1}{c|}{\textbf{13.39}} &
  \multicolumn{1}{c|}{\textbf{34.67}} &
  \textbf{0.17} \\ \cline{2-8} 
 &
  \textbf{100\%} &
  \multicolumn{1}{c|}{\textbf{42.16}} &
  \multicolumn{1}{c|}{57.87} &
  \textbf{0.42} &
  \multicolumn{1}{c|}{\textbf{23.40}} &
  \multicolumn{1}{c|}{\textbf{42.90}} &
  \textbf{0.26} \\ \hline
\multirow{5}{*}{Gemma 2-2B} &
  \textbf{0\%} &
  \multicolumn{1}{c|}{5.82} &
  \multicolumn{1}{c|}{22.71} &
  0.10 &
  \multicolumn{1}{c|}{4.61} &
  \multicolumn{1}{c|}{20.78} &
  \textbf{0.07} \\ \cline{2-8} 
 &
  \textbf{1\%} &
  \multicolumn{1}{c|}{14.36} &
  \multicolumn{1}{c|}{35.06} &
  0.21 &
  \multicolumn{1}{c|}{9.01} &
  \multicolumn{1}{c|}{27.99} &
  \textbf{0.15} \\ \cline{2-8} 
 &
  \textbf{10\%} &
  \multicolumn{1}{c|}{30.58} &
  \multicolumn{1}{c|}{\textbf{49.32}} &
  0.34 &
  \multicolumn{1}{c|}{15.99} &
  \multicolumn{1}{c|}{36.12} &
  0.22 \\ \cline{2-8} 
 &
  \textbf{50\%} &
  \multicolumn{1}{c|}{41.32} &
  \multicolumn{1}{c|}{\textbf{57.18}} &
  \textbf{0,42} &
  \multicolumn{1}{c|}{22.30} &
  \multicolumn{1}{c|}{\textbf{41.69}} &
  \textbf{0.27} \\ \cline{2-8} 
 &
  \textbf{100\%} &
  \multicolumn{1}{c|}{44.12} &
  \multicolumn{1}{c|}{59.10} &
  \textbf{0.45} &
  \multicolumn{1}{c|}{\textbf{23.50}} &
  \multicolumn{1}{c|}{42.49} &
  \textbf{0.28} \\ \hline
\multirow{2}{*}{\textbf{\begin{tabular}[c]{@{}c@{}}Luxembourgish\\ to  English\end{tabular}}} &
  \multicolumn{1}{l|}{\multirow{2}{*}{}} &
  \multicolumn{3}{c|}{\textbf{Val 300}} &
  \multicolumn{3}{c|}{\textbf{FLORE 200}} \\ \cline{3-8} 
 &
  \multicolumn{1}{l|}{} &
  \multicolumn{1}{c|}{\textbf{SPBLEU}} &
  \multicolumn{1}{c|}{\textbf{CharF++}} &
  \textbf{Jaccard} &
  \multicolumn{1}{c|}{\textbf{SPBLEU}} &
  \multicolumn{1}{c|}{\textbf{CharF++}} &
  \textbf{Jaccard} \\ \hline
\multirow{5}{*}{Llama 3.2 -3B} &
  \textbf{base Model} &
  \multicolumn{1}{c|}{26.31} &
  \multicolumn{1}{c|}{45.98} &
  0.33 &
  \multicolumn{1}{c|}{17.62} &
  \multicolumn{1}{c|}{36.79} &
  0.26 \\ \cline{2-8} 
 &
  \textbf{1\%} &
  \multicolumn{1}{c|}{34.18} &
  \multicolumn{1}{c|}{54.63} &
  0.4 &
  \multicolumn{1}{c|}{22.68} &
  \multicolumn{1}{c|}{45.98} &
  0.32 \\ \cline{2-8} 
 &
  \textbf{10\%} &
  \multicolumn{1}{c|}{43.28} &
  \multicolumn{1}{c|}{61.86} &
  0.48 &
  \multicolumn{1}{c|}{26.11} &
  \multicolumn{1}{c|}{50.51} &
  0.36 \\ \cline{2-8} 
 &
  \textbf{50\%} &
  \multicolumn{1}{c|}{\textbf{49.60}} &
  \multicolumn{1}{c|}{\textbf{67.15}} &
  \textbf{0.53} &
  \multicolumn{1}{c|}{\textbf{29.18}} &
  \multicolumn{1}{c|}{\textbf{54.35}} &
  \textbf{0.39} \\ \cline{2-8} 
 &
  \textbf{100\%} &
  \multicolumn{1}{c|}{57.88} &
  \multicolumn{1}{c|}{73.46} &
  0.60 &
  \multicolumn{1}{c|}{32.56} &
  \multicolumn{1}{c|}{57.60} &
  0.41 \\ \hline
\multirow{5}{*}{Gemma 2-2B} &
  \textbf{base Model} &
  \multicolumn{1}{c|}{27.11} &
  \multicolumn{1}{c|}{47.44} &
  0.34 &
  \multicolumn{1}{c|}{14.99} &
  \multicolumn{1}{c|}{37.77} &
  0.26 \\ \cline{2-8} 
 &
  \textbf{1\%} &
  \multicolumn{1}{c|}{43.00} &
  \multicolumn{1}{c|}{59.80} &
  0.47 &
  \multicolumn{1}{c|}{29.25} &
  \multicolumn{1}{c|}{49.15} &
  \textbf{0.38} \\ \cline{2-8} 
 &
  \textbf{10\%} &
  \multicolumn{1}{c|}{54.41} &
  \multicolumn{1}{c|}{68.86} &
  0.58 &
  \multicolumn{1}{c|}{36.14} &
  \multicolumn{1}{c|}{55.67} &
  0.45 \\ \cline{2-8} 
 &
  \textbf{50\%} &
  \multicolumn{1}{c|}{61.26} &
  \multicolumn{1}{c|}{\textbf{73.91}} &
  \textbf{0.64} &
  \multicolumn{1}{c|}{41.06} &
  \multicolumn{1}{c|}{\textbf{59.94}} &
  \textbf{0.49} \\ \cline{2-8} 
 &
  \textbf{100\%} &
  \multicolumn{1}{c|}{62.75} &
  \multicolumn{1}{c|}{75.13} &
  0.65 &
  \multicolumn{1}{c|}{\textbf{42.73}} &
  \multicolumn{1}{c|}{61.25} &
  \textbf{0.51} \\ \hline
\end{tabular}%
}
\end{table*}

\begin{table}[htbp]
\centering
\caption{Performance testing after SFT on Corresponding Validation Dataset (\#1000 samples)}
\label{tab:general_langauge_results}
\begin{adjustbox}{max width=\columnwidth}
\begin{tabular}{llcccc}
\hline
\textbf{Language Pair} & \textbf{Methods} & \textbf{SPBLEU} & \textbf{CharF++} & \textbf{Jaccard} & \textbf{LLMaaJ} \\
\hline
As-En & BM & 8.75 & 22.72 & 0.16 & 0.64 \\
      & DN & 9.00 & 23.03 & 0.16 & 0.65 \\
      & DL & 8.87 & 23.04 & 0.16 & 0.59 \\
      & DG & 9.43 & 23.69 & 0.16 & 0.62 \\
\hline
En-As & BM & 2.27 & 10.84 & 0.03 & 0.37 \\
      & DN & 8.75 & 22.72 & 0.16 & 0.64 \\
      & DL & 8.09 & 29.03 & 0.18 & 0.61 \\
      & DG & 8.07 & 29.23 & 0.18 & 0.65 \\
\hline
Kh-En & BM & 0.63 & 14.66 & 0.06 & 0.05 \\
      & DN & NA & NA & NA & NA \\
      & DL & 2.79 & 18.66 & 0.10 & 0.10 \\
      & DG & 4.81 & 23.43 & 0.14 & 0.30 \\
\hline
En-Kh & BM & 0.22 & 0.50 & 0.00 & 0.00 \\
      & DN & NA & NA & NA & NA \\
      & DL & 4.81 & 16.95 & 0.15 & 0.17 \\
      & DG & 11.58 & 29.19 & 0.23 & 0.51 \\
\hline
Uk-En & BM & 22.50 & 41.35 & 0.30 & 0.72 \\
      & DN & 25.34 & 44.06 & 0.33 & 0.77 \\
      & DL & 25.29 & 44.08 & 0.33 & 0.76 \\
      & DG & 24.81 & 43.76 & 0.32 & 0.78 \\
\hline
En-Uk & BM & 13.57 & 30.19 & 0.15 & 0.60 \\
      & DN & 17.87 & 34.83 & 0.18 & 0.70 \\
      & DL & 17.97 & 34.83 & 0.19 & 0.69 \\
      & DG & 18.10 & 34.97 & 0.19 & 0.72 \\
\hline
En-Lb & BM & 6.46 & 26.78 & 0.12 & 0.36 \\
      & DN & 37.98 & 55.41 & 0.37 & 0.82 \\
      & DL & 40.71 & 59.02 & 0.44 & 0.87 \\
      & DG & 44.58 & 59.73 & 0.45 & 0.87 \\
\hline
Lb-En & BM & 26.31 & 45.98 & 0.33 & 0.58 \\
      & DN & 42.78 & 59.33 & 0.48 & 0.82 \\
      & DL & 54.64 & 70.98 & 0.57 & 0.82 \\
      & DG & 59.88 & 74.97 & 0.63 & 0.90 \\
\hline
\end{tabular}
\end{adjustbox}
\end{table}

\section{Case Study}
\label{Appendix: Case Study Main section}

This section mainly presents several interesting text cases encountered during the pre-trained LLM generation process, which were identified through our manual quality checks.

\subsection{Uncontrollable output}
A notable translation issue arises from unnecessary explanatory notes appended by the model, which negatively affects automated evaluation metrics. Consider the following example, where the original English input was:

\begin{tcolorbox}[breakable]
"He added that ``they should not, however, be asked to take on obligations that go beyond their development stage, responsibility and capabilities.''"
\end{tcolorbox}

The model produced:

\begin{tcolorbox}[breakable]
"Dëi gouf och gesäftlech gesëtt datt ``si si och net an Obergrenzen vum ieren Entwéckelungsschatz, Responsabilität a Fähigkeeten uerecht bestallt.'' (Note: Please keep in mind that Luxembourgish is a complex and nuanced language, and translations may vary depending on regional dialects and contexts. This translation is provided in standard Luxembourgish)"
\end{tcolorbox}

The corresponding ground truth translation in Luxembourgish is:

\begin{tcolorbox}[breakable]
"Hien huet bäigefüügt, datt ``se awer net opgefuerdert sollte ginn, Verflichtungen ze iwwerhuelen, déi iwwer hiren Entwécklungsniveau, hir Verantwortung a Fäegkeeten erausginn.''"
\end{tcolorbox}

The inclusion of this extraneous note, likely originating from instruction-tuned general translation settings or cautious model outputs designed for conversational contexts, significantly impacts BLEU and related metrics. Such notes introduce irrelevant lexical content and length discrepancies, lowering similarity scores and thereby inaccurately reflecting the model's genuine translation ability.

To mitigate this issue, explicit prompt engineering or fine-tuning with specific stopping criteria and clearer output formatting instructions is essential. This strategy helps ensure that model-generated translations remain concise, relevant, and aligned closely with evaluation metric expectations.
\subsection{Repetition Problem}
\label{apd:repetionproblems}

During pre-training, we frequently encountered the issue of model hallucination, characterized by repetitive generation. This problem was notably less severe in the Gemma model compared to the Llama model. The repetitive generation primarily occurred in the English-to-Luxembourgish translation direction, largely due to the model's insufficient understanding of Luxembourgish. Consequently, the model frequently confused Luxembourgish with German, exacerbating this confusion further. 

Below is an example of input:

\begin{tcolorbox}[breakable]
Translate the following English input text into Luxembourgish. Do not include any additional information or unrelated content.
Input: The United States Strategic Command of the U.S. Department of Defense office is tracking the debris.
\end{tcolorbox}

Below is an illustrative output that presents repetitive generation:

\begin{tcolorbox}[breakable]
Here is the translation: D'United States Strategic Command vum ustrategeschen vum ustrategeschen Ufuerderungsbüro vum ustrategeschen Ufuerderungsbüro vum ustrategeschen Ufuerderungsbüro vum ustrategeschen Ufuerderungsbüro vum ustrategeschen Ufuerderungsbüro vum ustrategeschen ...
\end{tcolorbox}

We also observed that when distillation datasets contained a small amount of hallucinated data—such as infinitely repeating Luxembourgish content (particularly prominent in Llama 3.3 70B)—the subsequently fine-tuned models strongly inherited these hallucination patterns, resulting in substantial performance degradation. Therefore, maintaining high-quality data for fine-tuning is crucial. In comparison, data distilled from ChatGPT rarely exhibited infinite repetition issues, enabling the model to learn appropriate stopping criteria during generation. Additionally, we found that incorporating certain generation prompts during training, such as "Here is the translation: ", could partially mitigate the inherited infinite generation problems originating from distilled models of Llama.

\subsection{Hallucination to German}
\label{Appendix: Case Study Hallucination to german}

Given the English input sentence: 

\begin{tcolorbox}[breakable]
"At first glance, it looks slightly chaotic at the arrival, the children make a game of clapping hands with everyone who arrives, but that also contributes to the special charm of the event. The Widdebierg is not as high as one might assume from the logo, but it is still 386 meters, which is not nothing,"
\end{tcolorbox}

the model produced the following output:

\begin{tcolorbox}[breakable]
"Bei der Ankunft blickt man zunächst etwas chaotisch aus, die Kinder machen ein Spiel mit der Klatsch der Hände mit jedem, der kommt, aber das tragen auch zur besonderen Charme vun der Ävent vun der Éieren. Dëi Widdebiergëstcher Ääschten, well dass méi méi vun der Logo, mä si méi och 386 Méter, wéi déi éis gët nach Léck, wat déiweis méi och net."
\end{tcolorbox}

In contrast, the Luxembourgish ground truth is:

\begin{tcolorbox}[breakable]
"Op den éischte Bléck gesäit et op der Arrivée liicht chaotesch aus, d'Kanner maache sech e Spaass draus, jidderengem, deen ukënnt, an d’Hand ze klatschen, mä och dat dréit zum spezielle Charme vun der Manifestatioun bäi. De Widdebierg ass wuel net esou héich wéi een dat um Logo kéint unhuelen, mä ëmmerhi sinn et 386 Meter, dat ass net grad näischt."
\end{tcolorbox}

This incorrect translation output primarily results from excessive usage of German vocabulary rather than proper Luxembourgish expressions. This phenomenon likely arises due to several factors:

\begin{itemize}
    \item \textbf{Data Sparsity and Language Proximity:} Luxembourgish and German share considerable lexical and syntactic similarities. In conditions of limited Luxembourgish-specific training data, the model might unintentionally rely heavily on its knowledge of German, leading to significant linguistic interference.
    \item \textbf{Pretraining Corpus Bias:} The predominance of German texts over Luxembourgish in multilingual pretraining datasets likely reinforces German lexical and structural patterns, especially under resource-constrained fine-tuning conditions.
    \item \textbf{Limited Distinctive Training Examples:} Insufficient distinct Luxembourgish examples during fine-tuning might not effectively guide the model away from Germanic lexical choices, resulting in mixed-language outputs or incorrect lexical selections.
\end{itemize}

Addressing this issue effectively requires either extensive additional training data or targeted linguistic resources explicitly designed to emphasize lexical and grammatical distinctions between closely related languages such as Luxembourgish and German.

\section{Prompt Design for LLM}

\subsection{Prompt for LLM-as-a-Judge}
\label{apd:laajpmt}

For the prompt, we mainly adopt the previous legal translation prompt structure \cite{niklaus2025swiltrabenchswisslegaltranslation} but customize it simply for only the transation needs without any domain emphasis specification. In this paper, we primarily employ google/gemma-3-27b-it as the evaluation model to assess translation quality, given its strong instruction-following capabilities and competitive performance among open-weight LLMs. For efficient model inference, we adopt SGLang as the serving framework, which enables streamlined deployment and low-latency response for both evaluation and generation tasks.

\begin{tcolorbox}[breakable]

Your task is to assess the accuracy, clarity, and fidelity of the model's translation to the golden translation. \newline

You will be provided the golden translation, and the model's translation. Your task is to judge how correct the model's translation is based on the golden translation, and then give a correctness score. The correctness score should be one of the below numbers: 0.0 (totally wrong), 0.1, 0.2, 0.3, 0.4, 0.5, 0.6, 0.7, 0.8, 0.9, or 1.0 (totally right). You should give the correctness score directly. The correctness score must strictly follow this format: "[[score]]", e.g., "The correctness score: [[0.5]]. \newline
Golden Translation: \textbf{\{Golden Translation\}}
 \newline

Model Translation: \textbf{\{Model's Translation\}}
 \newline
\end{tcolorbox}

\subsection{Prompt for SFT}
\label{apd:proptforsft}

We primarily adopt the classical SFT approach, where the model is trained to predict the next token by minimizing the cross-entropy loss. Consequently, training data typically consist of input-output pairs, such as question-answer or instruction-response formats. The input is usually referred to as the prompt and the output as the answer. During training, the prompt and answer are concatenated and fed into the model, with the objective of guiding the model to generate the answer portion. In this work, we employ the following training template.

\begin{tcolorbox}[breakable]
Below is an instruction that describes a task, paired with an input that provides further context. Write a response that appropriately completes the request. \newline

\#\#\# Instruction: \newline
Translate the following English input text into Luxembourgish. Do not include any additional information or unrelated content. \newline
\newline
\newline
\#\#\#  Input:\newline
\textbf{\{The sentence to be translated\}}\newline
\newline

\#\#\#  Response:

\textbf{\{The translated sentence\}}\newline

\end{tcolorbox}

\end{document}